\documentclass[pdflatex,sn-mathphys]{sn-jnl}
\usepackage{subcaption}
\usepackage{lscape}
\usepackage{url}

\jyear{2023}%

\theoremstyle{thmstyleone}%

\theoremstyle{thmstyletwo}%

\theoremstyle{thmstylethree}%
\newtheorem{definition}{Definition}%

\begin{document}

\title[ ]{Interpretable and Explainable Machine Learning Methods for Predictive Process Monitoring: A Systematic Literature Review}

\author*[1,2]{\fnm{Nijat} \sur{Mehdiyev}}\email{nijat.mehdiyev@dfki.de}
\author[1,2]{\fnm{Maxim} \sur{Majlatow}}\email{maxim.majlatow@dfki.de}
\author[1,2]{\fnm{Peter} \sur{Fettke}}\email{peter.fettke@dfki.de}
\affil[1]{\orgname{German Research Center for Artificial Intelligence (DFKI)}, \orgaddress{\street{Campus D 3.2}, \city{Saarbrücken}, \postcode{66123}, \state{Saarland}, \country{Germany}}}
\affil[2]{\orgname{Saarland University}, \orgaddress{\street{Campus D 3.2}, \city{Saarbrücken}, \postcode{66123}, \state{Saarland}, \country{Germany}}}

\abstract{This paper presents a systematic literature review (SLR) on the explainability and interpretability of machine learning (ML) models within the context of predictive process mining, using the PRISMA framework. Given the rapid advancement of artificial intelligence (AI) and ML systems, understanding the "black-box" nature of these technologies has become increasingly critical. Focusing specifically on the domain of process mining, this paper delves into the challenges of interpreting ML models trained with complex business process data. We differentiate between intrinsically interpretable models and those that require post-hoc explanation techniques, providing a comprehensive overview of the current methodologies and their applications across various application domains. Through a rigorous bibliographic analysis, this research offers a detailed synthesis of the state of explainability and interpretability in predictive process mining, identifying key trends, challenges, and future directions. Our findings aim to equip researchers and practitioners with a deeper understanding of how to develop and implement more trustworthy, transparent, and effective intelligent systems for predictive process analytics.}
\keywords{Explainable Artificial Ingelligence (XAI), Interpretable Machine Learning, Predictive Process Monitoring, Process Mining}

\maketitle
\clearpage

\NewDocumentCommand{\rot}{O{90} O{0.25em} m}{\makebox[#2][c]{\rotatebox{#1}{#3}}}%
\label{introduction}
\section{Introduction}
The concepts of explainability and interpretability are crucial in the research domain of intelligent information systems, which is undergoing rapid development. These are multifaceted notions aiming to illuminate the inner workings of the adopted artificial intelligence (AI) and machine learning (ML) systems \cite{BARREDOARRIETA202082}. Moreover, the objective is to make the complex decisions and operations of these systems understandable and justifiable to human users. The significance of these concepts in intelligent systems is not a new research area; it has been a focal point of academic investigation for several decades \cite{wick1992reconstructive, gregor1999explanations}. The recent surge in advanced ML techniques has intensified the urgency of this exploration \cite{gunning2019xai}. The novel and efficient ML methods are often referred to as "black boxes" due to their complexity and opacity, rendering their operations non-transparent to users and stakeholders \cite{guidotti2018survey}.

In a conceptual sense, an explanation serves the purpose of bridging the gap between the high-level, frequently abstract operations of AI models and the practical, tangible understanding that users require \cite{lipton}. From a theoretical standpoint, this entails breaking down the layers of complicated algorithms in order to translate their obscure data patterns and decision pathways into a format that is understandable and significant to human beings \cite{fleisher2022understanding}. Not only is the opacity in data-driven solutions a theoretical concern, but it also has implications in the real world, particularly when it comes to algorithm aversion. According to Dietvorst et al. (2015), human users have a tendency to mistrust or even reject systems that they are unable to comprehend or that they believe they are unable to gain control over \cite{dietvorst2015algorithm}. As a result, the search for explainability or interpretability is not merely an academic pursuit but rather a practical necessity. 

Recently, there's been a notable trend toward comprehensive reviews  \cite{BARREDOARRIETA202082, guidotti2018survey, angelov2021explainable} and in-depth studies on AI explainability. The studies from the second category dissect the subject's complexities within targeted domains, including healthcare \cite{nazar2021systematic}, industry \cite{ahmed2022artificial}, law  \cite{vale2022explainable}, energy systems \cite{machlev2022explainable}, insurance \cite{owens2022explainable}, finance \cite{chen2023explainable}, education \cite{farrow2023possibilities}, material science \cite{zhong2017intelligent} etc., or they often focus specifically on certain data types, such as geospatial data \cite{roussel2023geospatial}, time series data  \cite{theissler2022explainable}, image data \cite{kamakshi2023explainable}, and text data \cite{gurrapu2023rationalization}. Even though these studies, when taken as a whole, offer a comprehensive overview of the methods that are currently in use and derive valuable design propositions, they frequently remain confined within their respective domains and data types or are excessively general.

Our study, on the other hand, focuses on interpretable and explainable ML within the specific research domain of process mining. Process mining is a field that lies at the intersection of data science and business process management (BPM) \cite{van2012_process_manifesto}. It is a family of techniques that makes use of event log data from process-aware information systems (PAIS) to deduce insights about the execution of processes in different application domains. Because of the inherent complexity of the sequential process data, which is also characterized by intricate branching and activities that may occur concurrently, the task of prediction and subsequent explanation represents a particularly difficult challenge \cite{mehdiyev2021explainable}. Over the course of the past decade, there has been a substantial rise in the amount of research conducted on predictive process monitoring, a branch of process mining. Numerous reviews and surveys have been conducted to investigate various aspects of this field \cite{neu2022systematic, di2018predictive,Maggi2014}. More recently, there has been a shift toward making black-box models used in predictive process monitoring more explainable. Differnt studies that investigate different approaches to XAI have been produced due to the surge in research that has taken place. Despite these endeavors, a systematic and comprehensive analysis of these methods remains elusive, with only a brief research-in-progress paper offering relevant insights but suffering from limited scope \cite{stierle2021bringing}.

Our research effort aims to address the discrepancy by conducting a systematic literature review (SLR) of interpretable and explainable ML methods for predictive process monitoring. It is essential to differentiate between intrinsically interpretable models, which are understandable by their very nature, and black-box models, which are more complex and require post-hoc explanation techniques \cite{guidotti2018survey}. In this paper, we will dissect and synthesize these methodologies, providing a coherent, comprehensible, and academically rigorous perspective on the current state of explainability and interpretability in predictive process mining and the future directions it will take. This study aims to provide researchers and practitioners with conceptual, theoretical, and practical implications for developing and implementing intelligent systems that are more trustworthy, transparent, and efficient. This will be accomplished through the bibliographic analysis that will be presented in this study.
\hfill

\label{background}
\section{Background}
In this section, we will delve into the fundamental aspects of process mining and predictive process monitoring. It is organized into comprehensive subsections, the first of which begins with a description of the primary ideas and formal definitions that are essential to the components of process mining. After that, we move on to the topic of predictive process mining, where we go into detail about the crucial components of the data pipeline as well as the various types of problem areas that are associated with this field. Following this, we delve into the detailed differentiation between interpretable and explainable ML, thereby furnishing a fundamental understanding of these notions. This is supplemented by formal definitions and discussions of the relevant methods that are utilized within the field. This systematic approach guarantees a comprehensive and unambiguous presentation of the essential background, thereby laying the groundwork for a more in-depth investigation into the intersection of ML, interpretability/explainability, and predictive process monitoring.

\subsection{Predictive Process Monitoring}
Over the past decade, there has been an increased interest in predictive process monitoring, which is a specific field within process mining \cite{van2012aprocess, breuker2016comprehensible}. The increased interest in this area can be mostly attributed to the competitive nature of industries in which process excellence is the main differentiator and the advancements in high-performing ML models \cite{Mehdiyev2020}. This research area focuses on predicting the future states of ongoing process executions \cite{evermann2017_next_event}. For this purpose, the digital footprints of previous process instances stored in an event log are used \cite{di2018predictive}. An event log consists of traces that record events in a sequential manner, providing an outline of the workflow for relevant procedures. We introduce the following established definitions to facilitate a comprehensive understanding and to ensure a unified basis for subsequent discussions.

\begin{definition}[\textbf{Event}]\label{def_event}
An $event$ is denoted by the tuple $e = (a, c, t_{start}, t_{complete},$ \\
$ v_{1}, \ldots, v_{n})$, where $a \in \mathcal{A}$ is a categorical variable denoting the process activity, $c \in \mathcal{C}$ is a categorical variable signifying the unique identifier for the trace, also called \textit{case ID}, $t_{start}\in \mathcal{T}_{start}$ and $t_{complete}\in \mathcal{T}_{complete}$ represent the event's commencement and completion timestamp (utilizing an epoch time representation like Unix) respectively, and $v_{1}, \ldots, v_{n}$ denoting the event-specific attributes, where $\forall  1 \leq i \leq n : v_{i} \in \mathcal{V}_{i}$ denote the domain of the $i^{th}$ attribute. Consequently, these variables create a multi-dimensional space for the universe of events $\mathcal{E}$.   
\end{definition}

In essence, an event in the context of predictive process monitoring is a multi-faceted entity characterized by its activity type, its association with a specific process trace, its start and completion times, and any additional attributes that may be relevant. These elements collectively define a multi-dimensional space $\mathcal{E}$ which can be thought of as the set of all possible events that could occur in the system under study.

The exemplary Table \ref{Sample_Event_Log}, derived from a manufacturing scenario, depicts an event in each row, with the first event being characterized by its \textit{Activity} "Plasma Welding", its \textit{Start Time} "2019-04-18 06:26:47", its \textit{End Time} "2019-04-18 09:51:25", the resource (\textit{Worker ID}), the \textit{Processing Time} "03:24:38" as well as other variables. Based on Definition \ref{def_event} we now define traces and partial traces:

\begin{table}[t]
\centering
\caption{Process Event Log Sample} \label{Sample_Event_Log}
\begin{tabular}{|c|c|c|c|c|c|c|}
\hline
\textbf{Case}   & \textbf{}         & \textbf{Start}    & \textbf{End}  & \textbf{Worker}   & \textbf{}         & \textbf{Processing} \\
\textbf{ID}     & \textbf{Activity} & \textbf{Time}     & \textbf{Time} & \textbf{ID}       & \textbf{...}      & \textbf{Time} \\
\hline\hline 
162384          & Plasma            & 2019-04-18        & 2019-04-18    & 409               & ...               & 03:24:38\\
                & Welding           & 06:26:47          & 09:51:25      &                   & & \\
\hline
162384          & Grinding          & 2019-04-18        & 2019-04-18    & 108               & ...               & 06:55:44\\
                & Weld. Seam        & 12:11:30          & 19:07:14      &                   & & \\
\hline
162384          & Dishing           & 2019-04-23        & 2019-04-23    & 150               &                   & 07:43:40\\
                & Press (\#2)       & 10:50:31          & 18:34:11      &                   & & \\
\hline
162384          & Beading           & 2019-04-24        & 2019-04-24    & 726               &                   & 09:37:32\\
                &                   & 10:20:13          & 19:57:45      &                   & & \\
\hline
162384          & X-Ray             & 2019-04-25        & 2019-04-25    & 703               &                   & 00:02:09\\
                & Examination       & 10:26:23          & 10:28:32      &                   & & \\
\hline
162384          & Edge              & 2019-04-26        & 2019-04-26    & 742               &                   & 03:41:49\\
                & Deburring         & 09:08:38          & 12:50:27      &                   & & \\
\hline
...             & ...               & ...               & ...           & ...               & ... & \\
\hline
177566          & 3D Micro-         & 2021-06-21        & 2021-06-21	& 139               &                   & 03:21:59\\
                & step              & 07:04:38          & 10:26:37      &                   & ... & \\
\hline
177566          & Plasma            & 2021-06-22        & 2021-06-22    & 409               & ...               & 04:24:28\\
                & Welding           & 08:26:47          & 12:51:05      &                   & & \\
\hline
177566          & Grinding          & 2021-06-22        & 2021-06-22    & 108               & ...               & 04:25:40\\
                & Weld. Seam        & 14:41:30          & 19:07:10      &                   & & \\
\hline
177566          & Surface           & 2021-06-23        & 2021-06-23    & 108               & ...               & 03:25:40\\
                & Polishing         & 09:38:38          & 13:00:27      &                   & & \\
\hline
...             & ...               & ...               & ...           & ...               & ...               & ...\\
\hline
\end{tabular}
Exemplary event log, depicting the trace identifier (\textit{Case ID}), timestamps for \textit{Start Time} and \textit{End Time}, the executed \textit{Activity}, the executing resource (\textit{Worker ID}), as well as a label (\textit{Processing Time}).
\end{table}

\begin{definition}[\textbf{Trace, Partial Trace, Prefix and Suffix}]
A $trace$ $\sigma \in \mathcal{E}^{*}$ is a finite sequence of unique events $\sigma =\langle e_{1}, e_{2}, \ldots, e_{{\lvert \sigma \rvert }}\rangle$, with $\lvert \sigma \rvert$ denoting the amount of events in the trace, also called \textit{trace length}, ordered chronologically and pertaining to a shared trace identifier $c \in \mathcal{C}$, also called \textit{case ID}. We denote the set of all possible traces by \( \mathcal{S} \subseteq \mathcal{E}^{*} \), with each trace \( \sigma \in \mathcal{S} \) belonging to this universe.\\ 
\noindent A \textit{partial trace} is a subsequence \( \sigma' = \langle e_{i_1}, e_{i_2}, \ldots, e_{i_k} \rangle \) of a given trace \( \sigma \), where \( 1 \leq i_1 < i_2 < \ldots < i_k \leq \lvert \sigma \rvert \) and \( 1 \leq k < \lvert \sigma \rvert \). A partial trace also shares the same unique identifier \( c \in \mathcal{C} \) as its parent trace \( \sigma \). The set of all possible partial traces derived from \( \sigma \) is denoted by \( \mathcal{S}_{\sigma'} \). \noindent The \textit{prefix} and \textit{suffix} denote specific types of partial traces, yielded by employing the $hd^{i}(\sigma_{c})$ and $tl^{i}(\sigma_{c})$ functions, respectively. This is realized by employing a selection operator (.): $\sigma(i)=\sigma_{i}, \forall i \in \left[1,{\lvert \sigma \rvert }\right] \subset \mathbb{N}$, such that $hd^{i}(\sigma)=\langle e_{1}, e_{2}, \ldots, e_{\min (i, \lvert \sigma \rvert)}\rangle$ and $tl^{i}(\sigma)=\langle e_{w}, e_{w+1}, \ldots, e_{\lvert \sigma \rvert}\rangle$, where $w=\max (1,\lvert \sigma \rvert-i+1)$.
\end{definition}

\noindent In Table \ref{Sample_Event_Log}, two traces are depicted withe the \textit{Case ID}s "162374" and "177566". The first trace starts with "Plasma Welding" and concludes with "Edge Deburring", while the second trace is initiated with "3D Microstep" and terminated after "Surface Polishing", with the events pertaining to a trace following a chronological order.

\begin{definition}[\textbf{Event Log}]\label{def_event_log}
An \textit{event log} is denoted by the set \(Log\), where \({Log} = \{ \sigma_1, \sigma_2, \ldots, \sigma_n \} \) and \( \sigma_i \in \mathcal{S} \) for \( 1 \leq i \leq n\), $n \in \mathbb{N^+}$. Each \( \sigma_i \) is a trace as previously defined. The event log \(Log\) is a collection of traces that may or may not share the same unique identifiers \( c \in \mathcal{C} \).
\end{definition}

\noindent Based on Definition \ref{def_event_log}, Table \ref{Sample_Event_Log} represents an excerpt from an event log. Such event logs can be utilized to extract features and labels, which can then be leveraged for the construction of predictive models:

\begin{definition}[\textbf{Feature Extraction}]
Feature extraction is a mapping function denoted by \( \phi: \mathcal{E} \cup \mathcal{S} \rightarrow \mathcal{X} \), where \( \mathcal{E} \) is the set of all possible events, \( \mathcal{S} \) is the set of all possible traces, and \( \mathcal{X} \) is the feature space. Given an event \( e \in \mathcal{E} \) or a trace \( \sigma \in \mathcal{S} \), the function \( \phi \) transforms it into a feature vector \( x \in \mathcal{X} \). 
\noindent For event-level feature extraction, \( \phi_{\text{event}}: \mathcal{E} \rightarrow \mathcal{X}_{\text{event}} \) maps each event \( e \) to a feature vector \( x_{\text{event}} \) in the event-level feature space \( \mathcal{X}_{\text{event}} \), while for trace-level feature extraction, \( \phi_{\text{trace}}: \mathcal{S} \rightarrow \mathcal{X}_{\text{trace}} \) maps each trace \( \sigma \) to a feature vector \( x_{\text{trace}} \) in the trace-level feature space \( \mathcal{X}_{\text{trace}} \).
\end{definition}

\begin{definition}[\textbf{Labeling}]
Let \( \mathcal{Y} \) be the set of all possible response variable values. For a non-empty trace \( \sigma \neq \langle \rangle \) such that \( \sigma \in \mathcal{S} \), and \( \mathcal{S} \subseteq \mathcal{E}^{*} \), the labeling function \( \textit{resp}_{event}: \mathcal{E} \times \mathcal{S} \rightarrow \mathcal{Y} \), \( \textit{resp}(e, \sigma) = y \) maps an event \( e \) within the trace \( \sigma \) to its respective response variable value \( y \in \mathcal{Y} \), and is defined for all \( e \in \sigma \) and \( \sigma \in \mathcal{S} \). The labeling function \( \textit{resp}_{trace}: \mathcal{S} \rightarrow \mathcal{Y} \), \( \textit{resp}(\sigma) = y \) maps a trace \( \sigma \) to its respective response variable value \( y \in \mathcal{Y} \), and is defined for all \( \sigma \in \mathcal{S} \).
\end{definition}

\noindent The concepts of feature extraction and labeling serve as a mechanisms to associate specific attributes or outcomes with individual events within a trace. By mapping each event or trace to a response variable, the labeling function facilitates the transformation of raw event data into a format amenable to analytical or ML methods. This enables researchers and practitioners to derive insights, make predictions or evaluate hypotheses based on the labeled data. The feature extraction and labeling functions thus acts as bridges between the raw, multi-dimensional event space and the target outcomes or attributes, thereby enriching a dataset for more advanced analyses. On the basis of previous definitions, we are now able to formalize the concept of supervised learning in the context of predictive process monitoring:

\begin{definition}[\textbf{Supervised Learning}]
Supervised learning is a paradigm in ML where a predictive model is constructed based on a labeled dataset. The dataset \( \mathcal{D} \) is generated from an event log \( \text{Log} \), feature extraction function \( \phi: \mathcal{E} \cup \mathcal{S} \rightarrow \mathcal{X} \), and a use-case-dependent labeling function \( \textit{resp}: \mathcal{E} \times \mathcal{S} \rightarrow \mathcal{Y} \) or \( \textit{resp}: \mathcal{S} \rightarrow \mathcal{Y} \). Each entry in \( \mathcal{D} \) is a tuple \( (x, y) \), where \( x \in \mathcal{X} \) is a feature vector and \( y \in \mathcal{Y} \) is the corresponding response variable.
\noindent The dataset \( \mathcal{D} \) is partitioned into training \( \mathcal{D}_{\text{train}} \), validation \( \mathcal{D}_{\text{val}} \), and testing \( \mathcal{D}_{\text{test}} \) subsets. A predictive model \( f: \mathcal{X} \rightarrow \mathcal{Y} \) is trained on \( \mathcal{D}_{\text{train}} \) by minimizing a loss function \( \mathcal{L}(f(x), y) \).
\noindent The validation set \( \mathcal{D}_{\text{val}} \) is utilized for hyperparameter tuning and to mitigate the risk of overfitting. The testing set \( \mathcal{D}_{\text{test}} \) is employed to evaluate the generalization performance of the model, providing an unbiased assessment of its predictive capabilities.
\end{definition}

\noindent It should be noted that supervised learning on the event level can be considered a special case of trace-level supervised learning, in that partial traces of length one are being employed. With a variety of predictive process monitoring application scenarios (see Figure \ref{PPM Types}), we provide definitions for predominant prediction tasks:

\begin{figure}[h]
\centering
\includegraphics[width=12cm]{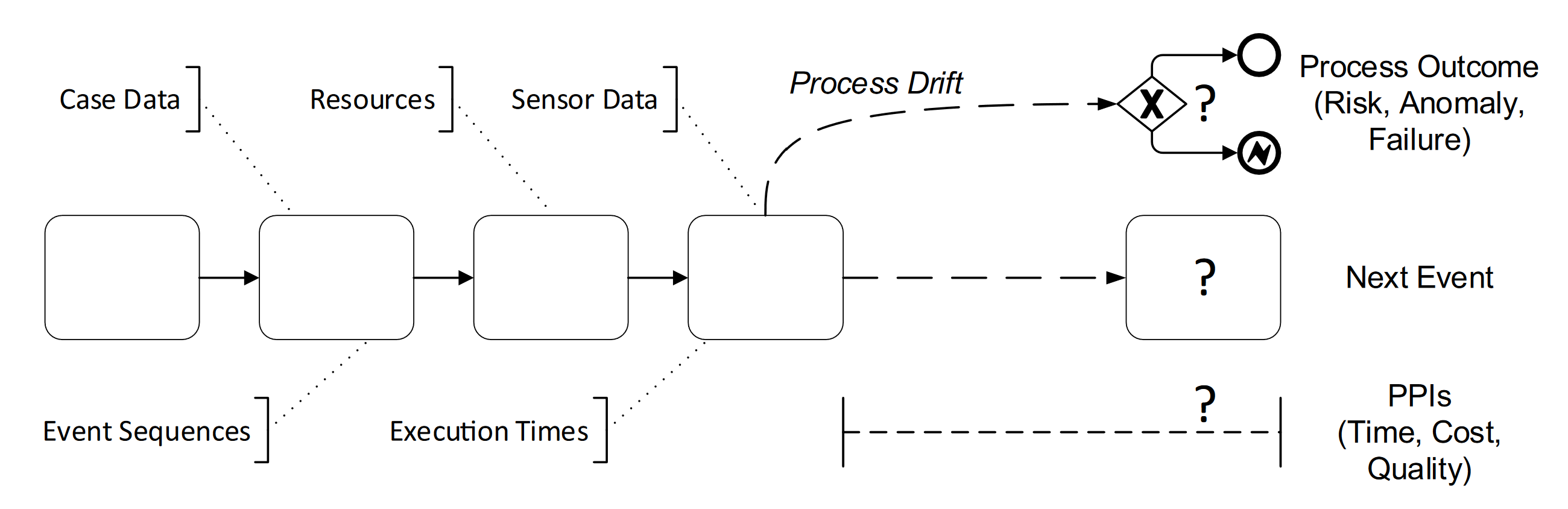}
\caption{Sources of input data accumulated in an event log and predictands of supervised learning \cite{DFKI_Smart_Lego_Factory}}
\label{PPM Types}
\end{figure}

\begin{definition}[\textbf{Process Outcome Prediction}]
Given a labeling function \( \textit{resp}_{\text{outcome}}: \mathcal{S} \rightarrow \mathcal{Y}_{\text{outcome}} \) mapping each (partial) trace \( \sigma \) to its final outcome \( y_{\text{outcome}} \), the predictive model \( f_{outcome}: \mathcal{X} \rightarrow \mathcal{Y}_{\text{outcome}} \) is constructed via supervised learning to approximate this function.
\end{definition}

\begin{definition}[\textbf{Next Event Prediction}]
Given a labeling function \( \textit{resp}_{\text{next}}: \mathcal{E} \times \mathcal{S} \rightarrow \mathcal{E}_{\text{next}} \)  mapping each event \( e \) within a trace \( \sigma \) to its subsequent event \( e_{\text{next}} \), the predictive model \( f_{next}: \mathcal{X} \rightarrow \mathcal{E}_{\text{next}} \) is constructed via supervised learning to approximate this function.
\end{definition}

\begin{definition}[\textbf{Process Performance Indicator (PPI) Prediction}]
Given a labeling function \( \textit{resp}_{\text{PPI}}: \mathcal{S} \rightarrow \mathcal{Y}_{\text{PPI}} \) mapping each (partial) trace \( \sigma \) to a performance metric \( y_{\text{PPI}} \), the predictive model \( f_{PPI}: \mathcal{X} \rightarrow \mathcal{Y}_{\text{PPI}} \) is constructed via supervised learning to approximate this function.
\end{definition}

Process data facilitates the development of predictive models that serve various objectives. These include the identification of the next likely activity \cite{evermann2017_next_event, Sindhgatta2020a}, the process outcome prediction \cite{Mehdiyev2020, Rizzi2020}, anomaly detection \cite{bohmer2020mining, pauwels2019detecting}, and reamining time prediction \cite{polato2014data, polato2018time}.  When it comes to developing accurate, reliable, and suitable models for the specific application context, the complexity and variability inherent in modern business processes may pose significant challenges. Additionally, the complexity of the models required to make such predictions is rising in tandem with the demand for more sophisticated estimations. Specifically, opaque models frequently achieve high predictive accuracy, which makes them appealing choices. Having said that, the complexity of these models presents a significant disadvantage, as they can be extremely difficult to grasp. For practical applications, where it is essential to comprehend the reasoning behind predictions to establish trust and make decisions, this is a significant limitation that must be considered \cite{marquez2017predictive, di2018predictive}. As a result, the development of models that strike a balance between accuracy and interpretability continues to be a significant challenge in the field of predictive process monitoring despite the fact that this area has tremendous potential.

\subsection{Interpretable and Explainable Artificial Intelligence}
\subsubsection{Foundations: Interpretability vs. Explainability}
The need for explainable and interpretable AI has been recognized for decades, with its importance underscored by the potential for bias and discrimination in decision-making \cite{confalonieri2021historical}. However, the criteria for a good explanation in this context remain unclear \cite{confalonieri2019makes}. To address this, Miller (2019) suggests drawing on research in philosophy, psychology, and cognitive science to understand how humans define, generate, and evaluate explanations \cite{miller2019explanation}. This approach is further supported by Emmert-Streib (2020), who emphasizes the importance of a reality-grounded perspective in the development of explainable AI \cite{emmert2020explainable}. 

Sokol (2021) defines explainability as a process of logical reasoning applied to transparent insights, interpreted under background knowledge, and placed within a specific context \cite{sokol2021explainability}. This understanding is further developed by Amgoud (2022), who introduces key axioms for explainers that provide reasons behind decisions, distinguishing between those that return sufficient reasons and those that provide necessary reasons \cite{amgoud2022axiomatic}. Hallé (2021) extends these concepts to the formal foundations of explainability for abstract functions, establishing explanation relationships for elementary functions and their compositions \cite{halle2021foundations}. On the other hand, Yang (2022) further delves into the psychological theory of explainability, proposing that humans interpret AI-generated explanations by comparing them to their own \cite{yang2022psychological}. Wicklund (2012) cautions against over-simplifications in psychological theories, showing the role of the "explainer" and the potential for bias theory formulation \cite{wicklund2012zero}. 

The field of organization sciences also offers a unique lens through which to understand the concept of AI explainability. Hafermalz (2021) highlights the need to consider the organizational perspective in generating explainability, posing key questions about the user, purpose, and location of explanations \cite{hafermalz2022please}. Ehsan (2021) further expands this by introducing the concept of Social Transparency (ST) in XAI, emphasizing the importance of incorporating the socio-organizational context into AI-mediated decision-making \cite{ehsan2021expanding}. Abedin (2021) adds a contingency theory framework to the discussion, identifying and managing the opposing effects of AI explainability, such as comprehensibility, conduct, confidentiality, completeness, and confidence in AI \cite{abedin2022managing}. These perspectives from different domains underscore the importance of understanding human neesd, the role of transparency and predictive power, and the need for user- and context-focused explanations in developing explainable AI systems.

Various studies highlight the distinction between interpretability and explainability, with the former focusing on contextualizing model output and the latter on describing the mechanism behind it. The distinctions between these two notions are subtle yet significant, and understanding them is crucial for the responsible development and deployment of AI systems  \cite{guidotti2018survey,BARREDOARRIETA202082}. Interpretable AI is fundamentally about the model's inherent transparency and the ability for its decisions to be directly understood by humans. It implies that the model's decision-making process is transparent and its workings can be comprehended without additional aids or explanations \cite{rudin2019stop}. For instance, decision trees are often cited as interpretable models because their decision-making process is straightforward and can be visualized and understood by examining the series of decisions leading to a conclusion. The demand for interpretability is often driven by the need for reliability, safety, and fairness in AI applications \cite{rudin2022interpretable}. Freitas (2014) provide a comprehensive framework for understanding interpretability, discussing its importance in providing assurances that models behave as expected and can be trusted, especially in high-stakes decisions \cite{freitas2014comprehensible}. The pursuit of interpretable AI aligns with the broader quest for simplicity and clarity in scientific models, as eloquently discussed by Carvalho et al. (2019), who argue that interpretable models facilitate verification, validation, and insights into the model's behavior \cite{carvalho2019machine}.

On the other hand, explainable AI is somewhat broader and pertains to the set of methods and techniques used to help human users comprehend and trust the output of ML models, especially those that are inherently complex and opaque, like neural networks \cite{doshi2017towards}. Explainability does not necessarily mean the model itself is simple or interpretable, but rather that there is an additional layer or method that helps to elucidate how the model arrived at its decisions \cite{guidotti2018survey}. This could involve post-hoc explanation techniques, which seek to approximate and explain the predictions of complex models \cite{gunning2019xai}. These methods are not without their critiques, as highlighted by Lipton (2018), who points out the often ambiguous nature of what constitutes an 'explanation' and calls for a more rigorous, theoretically grounded approaches \cite{lipton}.

The distinction between interpretability and explainability is crucial because it aligns with different needs and applications. Interpretable models are often preferred for high-stakes domains where understanding the decision-making process is as critical as the decision itself. Conversely, in domains where performance is paramount and complex models are necessary, explainable AI becomes indispensable. Therefore, a critical perspective in this discourse is the trade-off between performance and transparency. As models become more complex and potentially more accurate, they often become less interpretable. This trade-off is a fundamental tension in AI development and raises significant ethical and practical considerations. For instance, in a healthcare setting, a highly accurate but completely opaque model could make decisions that impact patient care without clinicians or patients understanding why, raising issues of trust and accountability \cite{rudin2019stop}. The trade-off between interpretability and performance is not merely a technical challenge but a fundamental issue that touches upon the epistemology of AI. Murdoch et al. (2019) provide a detailed discussion on the trade-offs between accuracy and interpretability, emphasizing the need for a balance that respects both the utility and the ethical implications of AI systems \cite{murdoch2019definitions}. Similarly, Burrell (2016) explores the sources of opacity in ML, discussing the inherent trade-offs and the sociotechnical nature of the problem \cite{burrell2016machine}.\\

\subsubsection{Formal Definitions of interpretable and explainable AI}

After delineating the differences between interpretability and explainability within AI, this section provides formal definitions of the methods from both categories. This is necessary as we aim to encompass all pertinent techniques from each category relevant to predictive process monitoring. 

\begin{definition}[\textbf{Intrinsically Interpretable Model}]
Let \( \mathcal{M} \) be the class of predictive models. A model \( f \in \mathcal{M} \) is termed an \textit{intrinsically interpretable model} if it possesses a humanly interpretable internal structure, denoted by \( \mathcal{I}(f) \), such that \( \mathcal{I}(f): \mathcal{X} \rightarrow \mathcal{Z} \), where \( \mathcal{Z} \) is the space of humanly interpretable representations.
\end{definition}

\noindent Considering a production process scenario where the objective is to predict the remaining time until case completion, an intrinsically interpretable approach might involve using a decision tree that makes its predictions based on a small set of easily interpretable features, such as the type of activity and the duration of the previous event. Because decision trees are inherently interpretable, the model satisfies the interpretability constraints \(\mathcal{I}(f)\) intrinsically.
Among approaches that are commonly considered intrinsically interpretable, Stierle et al. \cite{stierle2021bringing} differentiate between rule-based (for example (evolutionary) decision rules \cite{malioutov2017learning, marquez2017run}), regression-based (for example logistic regression \cite{Teinemaa2016}), tree-based (for example decision trees \cite{AllahBukhsh2019}) and probabilistic models (for example Bayesian networks \cite{dey2005bayesian}). Additionally, algorithmically transparent approaches like K-nearest-neighbors \cite{kumar2005using} as well as  generalized additive models \cite{coussement2010improved} are generally considered transparent as well \cite{BARREDOARRIETA202082}. Nonetheless, it is worth noting that these white-box models are often outperformed by more complex, opaque models in terms of predictive accuracy \cite{guidotti2018survey}.

\begin{definition}[\textbf{Black-Box Model}]
Let \( \mathcal{M} \) be the class of predictive models. A model \( f \in \mathcal{M} \) is termed a \textit{black-box model} if its internal structure is not readily humanly interpretable, denoted by \( \mathcal{I}(f) = \emptyset \).
\end{definition}
\noindent The characteristics of black-box models encompass a complexity in their behaviour and decision making processes which necessitate post-hoc explanations for understanding, with deep learning methods\cite{Mehdiyev2020, Sindhgatta2020a}, gradient boosting models \cite{Petsis2022} and random forests \cite{Verenich2016} being among the most prominent.

\begin{definition}[\textbf{Post-hoc Local Explanations}]
Let \( \mathcal{M} \) be the class of predictive models, and \( f \in \mathcal{M} \) be a specific model with predictive mapping \( f: \mathcal{X} \rightarrow \mathcal{Y} \). A local explanation is denoted by \( f_{\text{local}}: \mathcal{M} \times \mathcal{X} \times \mathcal{Y} \rightarrow \mathcal{Z}_{\text{local}} \), where \( \mathcal{Z}_{\text{local}} \) is the space of interpretable local representations. For a given instance \( (f, x, y) \in \mathcal{M} \times \mathcal{X} \times \mathcal{Y} \), \( f_{\text{local}}(f, x, y) \) elucidates the model's decision \( f(x) = y \) in the vicinity of \( x \). Model-agnostic local explanations can take any \( f \in \mathcal{M} \) as input, whereas model-specific local explanations are restricted to a subset \( \mathcal{M}_{\text{local}, f} \subset \mathcal{M} \).
\end{definition}
\noindent Prominent examples of local post-hoc explanations are Individual Conditional Expectation (ICE) Plots \cite{ICE_Plot_Example_Goldstein} for single instances, SHapley Additive exPlanations (SHAP) \cite{SHAP_Lundberg} or locally interpretable surrogate models like LIME \cite{LIME_Ribeiro}, which are model-agnostic approaches. Model-specific approaches finding use in deep neural networks are layer-wise relevance propagation \cite{montavon2019layer} and DeepLIFT \cite{shrikumar2017learning}. For tree-based models exhibiting a high complexity, Tree Shapley Additive Explanations (TreeSHAP) \cite{lundberg2020local} realizes a model-specific explanation techniques. 

\begin{definition}[\textbf{Post-hoc Global Explanations}]
Let \( \mathcal{M} \) be the class of predictive models, and \( f \in \mathcal{M} \) be a specific model with predictive mapping \( f: \mathcal{X} \rightarrow \mathcal{Y} \). A global explanation is denoted by \( f_{\text{global}}: \mathcal{M} \times \mathcal{X} \times \mathcal{Y} \rightarrow \mathcal{Z}_{\text{global}} \), where \( \mathcal{Z}_{\text{global}} \) is the space of interpretable global representations. The function \( f_{\text{global}}(f, \mathcal{X}, \mathcal{Y}) \) elucidates the model's overall decision-making mechanism across the entire domain \( \mathcal{X} \). Model-agnostic global explanations can take any \( f \in \mathcal{M} \) as input, whereas model-specific global explanations are restricted to a subset \( \mathcal{M}_{\text{global}, f} \subset \mathcal{M} \).
\end{definition}
\noindent Prominent examples of global, model-agnostic post-hoc explanations are Accumulated Local Effects (ALE) \cite{apley2020visualizing}, Decision Rules \cite{frank1998generating, malioutov2017learning}, Feature Importance \cite{fisher2019all}, Partial Dependence Plots (PDP) \cite{PDP_Friedman} (also in conjunction with ICE plots \cite{ICE_Plot_Example_Goldstein}) and global surrogate models like CART decision trees \cite{rutkowski2014cart}. 

\subsection{Related Surveys and Contribution}\label{related_work}
The field of predictive process monitoring has been the subject of numerous studies and SLRs, each contributing valuable insights into different aspects of this rapidly evolving domain. This section contrasts the focus and contributions of prominent related studies, particularly review articles with the distinctive elements of our study, particularly emphasizing our exploration of interpretable and explainable AI within predictive process monitoring (see Table \ref{survey_related})

Márquez-Chamorro et al. (2018)  \cite{marquez2017predictive}, Teinemaa et al. (2019)  \cite{teinemaa2019outcome}, and Di Francescomarino et al. (2018) \cite{di2018predictive}, Maggie et al. (2014) \cite{Maggi2014} have provided comprehensive overviews of predictive process monitoring tasks, computational methods, and their evaluations. They discuss various computational predictive methods, from statistical techniques to ML approaches, and provide valuable insights into the applications and performance of various models. While these studies offer a substantial understanding of predictive process monitoring, they do not focus explicitly on interpretability and explainability. At most, these studies include a discussion of some interpretable AI methods, but XAI approaches, particularly those going beyond inherent model transparency, are not addressed at all. Kubrak et al. (2022) \cite{kubrak2022prescriptive} delve into prescriptive process monitoring, incorporating elements of XAI and interpretable AI. However, their focus is predominantly on prescriptive analytics, and while they mention relevant XAI papers, they do not provide an extensive overview of studies in this area, leaving a gap for a more focused and detailed exploration.

Stierle et al. (2021) \cite{stierle2021bringing} stand out as one of the few studies aiming to provide a systematic review of XAI approaches specifically for predictive process monitoring. They categorize literature according to purpose, evaluation method, and model complexity, differentiating between intrinsically interpretable models and opaque models requiring post-hoc explanations. However, being a research-in-progress paper and considering the rapid advancements and proliferation of research in this field, the scope of their review is somewhat limited. Our study addresses this by providing a more comprehensive and up-to-date review of XAI in predictive process monitoring. Furthermore, while Mehdiyev and Fettke (2021)  \cite{mehdiyev2021explainable} and El-khawag et al. (2022) \cite{el2022xai} discuss the necessity of XAI for predictive process monitoring and propose frameworks for building relevant solutions, they do not provide an SLR. Their contributions are valuable in demonstrating applied examples and discussing frameworks, but they do not offer a broad overview of the field.

\begin{table}[h]
\begin{center} 
\begin{minipage}{330pt}
\setlength{\tabcolsep}{3pt}
\caption{Summary and categorisation of related work.}
\label{survey_related}
\begin{tabular}{c|cccccccc|c}
\toprule
& \multicolumn{8}{c}{Related Work} & \\

Characteristics &
\rot{Márquez-Chamorro et al. \cite{marquez2017predictive}}  & 
\rot{Teinemaa et al. \cite{teinemaa2019outcome}} & 
\rot{Di Francescomarino et al. \cite{di2018predictive} }  & 
\rot{Maggie et al. \cite{Maggi2014}} & 

\rot{Kubrak et al. \cite{kubrak2022prescriptive} }  & 
\rot{Stierle et al. \cite{stierle2021bringing}}  & 
\rot{Mehdiyev and Fettke \cite{mehdiyev2021explainable}} &
\rot{El-khawag et al. \cite{el2022xai}}  & 

\rot{\textbf{This Article}} \\
\hline \midrule
\textit{Is the primary emphasis of the article on interpretability} & 
& 
& 
& 
& 

& 
$\blacksquare$ & 
$\blacksquare$ & 
$\blacksquare$ & 

$\blacksquare$  
\\
\textit{or explainability?} & &&&&&&&&\\ \midrule

\textit{Does the article include interpretable AI methods} & 
$\blacksquare$ & 
$\blacksquare$ & 
$\blacksquare$ & 
$\blacksquare$ & 

$\blacksquare$ & 
$\blacksquare$ & 
& 
&

$\blacksquare$  
\\
\textit{for Predictive Process Monitoring?} & &&&&&&&&\\ \midrule

\textit{Does the article include explainable AI methods} & 
& 
& 
& 
& 

& 
$\blacksquare$ & 
$\blacksquare$ & 
$\blacksquare$ & 

$\blacksquare$  
\\
\textit{for Predictive Process Monitoring?} & &&&&&&&&\\ \midrule

\textit{Does the article discuss the evaluation of} & 
& 
& 
 & 
& 

& 
$\blacksquare$ & 
$\blacksquare$ & 
$\blacksquare$ & 

$\blacksquare$  
\\
\textit{interpretability or explainability?} & &&&&&&&&\\ \midrule

\textit{Is the article a completed systematic review of literature?} & 
$\blacksquare$ & 
$\blacksquare$ & 
$\blacksquare$ & 
$\blacksquare$ & 

$\blacksquare$ & 
& 
& 
& 

$\blacksquare$  
\\
\bottomrule
\end{tabular}
\end{minipage}
\end{center}
\end{table}

In contrast, our contribution lies in the systematic and focused exploration of interpretable and explainable AI in predictive process monitoring. We build on the foundation laid by previous surveys but go further by explicitly focusing on XAI approaches. Our study systematically collects and synthesizes the latest research, providing a nuanced understanding of the characteristics, capabilities, and limitations of various XAI methods. We aim to fill the gaps left by previous studies, offering a comprehensive review that not only maps the current landscape but also critically assesses methodologies, identifies research gaps, and provides clear, evidence-based recommendations for researchers and practitioners. Our SLR thus contributes to a more organized, centralized understanding of XAI in predictive process monitoring, supporting informed decision-making and guiding future research in this vital area.
\hfill

\section{Methodology}\label{sec3}

To ensure a thorough and methodical approach, we conducted an SLR in this study using the PRISMA (Preferred Reporting Items for Systematic Reviews and Meta-Analyses) framework \cite{page2021prisma}. With this methodology, we can provide a transparent and structured process for our review. It encompasses a number of important aspects that direct our research. 

At the outset, we present a justification of the rationale that grounds our research, unambiguously defining the necessity of the investigation as well as its significance in the present academic and practical setting. After this, we will proceed to provide an outline of our objectives, which are particular and measurable goals that we intend to accomplish through the use of this SLR. The subsequent phase is to identify information sources, which entails determining the databases and other repositories that will be used to search for literature pertinent to the topic. Our search strategy has been rigorously planned to include particular keywords and criteria, guaranteeing an extensive and targeted retrieval of desired studies. The preceding section provides an in-depth description of the selection process, which outlines the procedures for screening and selecting articles that satisfy our predetermined criteria. This leads to the eligibility criteria, which constitute the principles that are established for including or excluding studies.

The next step is to provide a description of the data collection process, which includes a detailed explanation of how we extract and manage the data from the selected studies, ensuring that it is reliable and consistent. In order to provide a comprehensive understanding of the findings, the synthesis methods section explains the techniques utilized to analyze and combine data from various academic research studies. At last, we will review the results of syntheses, which will provide a summary of the combined outcomes of all the included studies. Additionally, we will present the findings from individual studies in order to provide a comprehensive account of each relevant research contribution. Our methodology adheres to the highest standards of systematic review since we have diligently handled each of these items. This ensures that our research conclusions are built on a foundation that is robust, transparent, and reproducible.

\subsection{Rationale and Objectives}\label{rationale}
The rationale for carrying out this SLR is firmly grounded in the ever-evolving and fast-paced domain of interpretable and explainable AI. In recent years, there has been also a significant increase in the number of academic studies that concentrate on the implementation of pertinent methodologies and concepts for the purpose of predictive process monitoring. Nevertheless, the rapid proliferation of academic investigation, combined with a lack of comprehensive meta-analytical studies, has resulted in a fragmented landscape of knowledge. The absence of a systematic framework and cohesive integration of knowledge presents notable challenges for researchers and practitioners alike, rendering the synthesis and practical application of existing information a formidable task.

In order to adequately address this matter, it is imperative to undertake an SLR, which will yield a comprehensive and well-structured synopsis of the present state of knowledge and advancements in the field. Conducting a comprehensive review of the recently proposed methods in explainable predictive process monitoring will facilitate a more profound comprehension of their inherent characteristics, capabilities, and limitations. For researchers, this framework provides a comprehensive basis for discerning areas of research that require further investigation, enabling them to concentrate their endeavors and potentially make valuable contributions towards addressing these gaps. For professionals in the field, a systematic review holds immense value as it enables them to make more informed and discerning judgments regarding the techniques that are most appropriate for their particular contexts. This aspect assumes paramount importance in light of the multifaceted nature and intricacy of the discipline, which may prove overwhelming and arduous to navigate in the face of the incessant stream of novel research and advancements.

The primary objectives of this SLR are centered around the provision of a comprehensive and nuanced comprehension of the domain under investigation. Through a comprehensive analysis of the existing research landscape, rigorous evaluation of the employed methodologies, awareness of gaps, and the provision of unambiguous, evidence-based recommendations, the primary objective of this review is to augment the quality and reliability of research conducted within this field. The primary objective of this work is to enhance the decision-making process by providing individuals with a greater depth of information. Additionally, it aims to enhance the dissemination of knowledge and the sharing of best practices in the field of process analytics across multiple industries. Ultimately, the overarching goal is to make significant contributions toward advancing predictive modeling by fostering transparency, reliability, and effectiveness. The key goal of this initiative is for the SLR to function as a highly beneficial asset for both the scholarly community and professionals in the industry. Its purpose is to guide in navigating the intricate realm of interpretable and explainable AI while simultaneously promoting this field's overall progress and credibility.

\subsection{Information Sources, Search Strategy, Selection Process}\label{search_process}
We have explored various online databases including ACM Digital Library, AIS eLibrary, IEEE Xplore, Science Direct and SpringerLink to gather relevant publications. These databases, which include but are not limited to topic-specific literature, were searched via queries. 

The search queries are specified as follows: Each query includes one of the terms "business process prediction", "predictive process monitoring", "prescriptive process analytics", or "process mining" and are combined with either of the terms "expla*", "interpretab*" or "XAI" via the AND-operator, in order to narrow the results to domain-specific subjects. Where it was possible, the following query was used to yield any potentially relevant literature from a database: $Q_{comp}$= (expla* OR interpret* OR XAI) AND ("process mining" OR "business process prediction" OR "predictive process monitoring" OR "prescriptive process analytics"). The Symbol "*", as in "expla*", is being used as a wildcard if a database allowed the usage of wildcards. In databases that did not allow using wildcards, the terms "explanation", "explainable" and "explainability" were used instead of "expla*", as well as "interpretable" and "interpretability" instead of "interpret*".

Table \ref{tab_search_process_queries} presents a concise summary of the composition and usage of queries in case $Q_{comp}$ could not be processed by a database.

\begin{table}[h]
\begin{center} 
\begin{minipage}{270pt}
\caption{Summary of employed search queries for retrival of relevant literature.}
\label{tab_search_process_queries}%
\begin{tabular}{@{}lll@{}}
\toprule
Representation  & Search query  &   Used for\\
                &               &   querying databases\\
\midrule
$Q_{1}$ &	"business process prediction" & False\\
$Q_{2}$ &	"predictive process monitoring" & False\\
$Q_{3}$ &	"prescriptive process analytics" & False\\ 
$Q_{4}$ &	"process mining" & False\\
$Q_{5}$ &	"expla*" & False\\
$Q_{6}$ &	"interpretab*" & False\\
$Q_{7}$ &	"XAI" & False\\
$Q_{1,5}$ & $Q_{1}$ AND $Q_{5}$ & True\\
$Q_{1,6}$ & $Q_{1}$ AND $Q_{6}$ & True\\
$Q_{1,7}$ & $Q_{1}$ AND $Q_{7}$ & True\\
$Q_{2,5}$ & $Q_{2}$ AND $Q_{5}$ & True\\
$Q_{2,6}$ & $Q_{2}$ AND $Q_{6}$ & True\\
$Q_{2,7}$ & $Q_{2}$ AND $Q_{7}$ & True\\
$Q_{3,5}$ & $Q_{3}$ AND $Q_{5}$ & True\\
$Q_{3,6}$ & $Q_{3}$ AND $Q_{6}$ & True\\
$Q_{3,7}$ & $Q_{3}$ AND $Q_{7}$ & True\\
$Q_{4,5}$ & $Q_{4}$ AND $Q_{5}$ & True\\
$Q_{4,6}$ & $Q_{4}$ AND $Q_{6}$ & True\\
$Q_{4,7}$ & $Q_{4}$ AND $Q_{7}$ & True\\
$Q_{comp}$& ($Q_{1}$ OR $Q_{2}$ OR $Q_{3}$ OR $Q_{4}$) &  True\\
& AND ($Q_{5}$ OR $Q_{6}$ OR $Q_{7}$) &\\

\botrule
\end{tabular}
\end{minipage}
\end{center}
\end{table}

The inconsistencies between the search tools of each of the aforementioned databases make it challenging to conduct a systematic literature search using only the specified queries. In order to conduct an exhaustive search, the queries were applied to the title, keywords and complete text where it was possible:

\begin{itemize}
  \item For the ACM Digital Library, the "Search items from"-option was set to "The ACM Full-Text collection", the queries were searched within "Anywhere" (see "Search Within"-option). The filter "Research Article" was applied.
  \item For the AIS eLibrary, the queries were searched within "All Fields"
  \item For the IEEE Xplore, the queries were searched using the "Command Search"-tool
\end{itemize}

Following the database querying, the resulting literature was filtered using pre-defined criteria (for details, see Section \ref{eligibility}). Subsequently, a forward and backward search was conducted on the results to capture additional topic-relevant publications that could not be discovered by searching the databases directly, including relevant articles from the arXiv outlet as well.

\subsection{Eligibility Criteria}\label{eligibility}
The studies retrieved only through a systematic search may nevertheless provide outcomes that are not topic-specific for this systematic review, necessitating additional screening to meet research rigor. Therefore, inclusion and exclusion criteria for the literature are defined. The identified literature must satisfy all of the predefined inclusion criteria while also not meeting any of the exclusion criteria in order to be considered for inclusion. A comprehensive list of all inclusion and exclusion criteria can be found in Table \ref{tab_in_ex_criteria}.
 
\begin{table}[h]
\begin{center} 
\begin{minipage}{330pt}
\caption{Inclusion and exclusion criteria}
\label{tab_in_ex_criteria}
\begin{tabular}{@{}lll@{}}
Representation  & Criteria for  &   Description\\
\hline\midrule 
$IN_{1}$ &	Inclusion & Publication outlet is a peer-reviewed source,  \\
         &	         & e.g. journal, conference proceedings, etc. \\
$IN_{2}$ &	Inclusion & Publication addresses PPM tasks \\
$IN_{3}$ &	Inclusion & Publication incorporates XAI methodology \\
$IN_{4}$ &	Inclusion & Publication is written in English \\
& & \\ \midrule
$EX_{1}$ &	Exclusion & Publication outlet is not a peer-reviewed source \\
         &	         & and not identified by forward-/backward search \\
$EX_{2}$ &	Exclusion & Publication does not address PPM tasks \\
$EX_{3}$ &	Exclusion & Publication neither incorporates XAI methodology \\
         &	         & nor uses any interpretable methods\\
$EX_{4}$ &	Exclusion & Publication does not use an event log\\
$EX_{5}$ &	Exclusion & Publication is not written in English\\
\end{tabular}
\end{minipage}
\end{center}
\end{table}

These criteria were applied in the following manner: After querying a database, the title and abstract of each of the resulting publications were analyzed respectively with regards to the inclusion and exclusion. This represents first filtering step after the retrieval of literature. The next filtering step takes place by expanding the analysis from title and abstract to the full text of each publication that passed the first filtering step.
Based on the results of the second filtering step, a forward and backward search was conducted, which immediately applied filtering with the previously described inclusion and exclusion criteria.

\subsection{Data Collection Process and Synthesis Methods}\label{synthesis}

The primary phase of our data collection procedure entails the methodical extraction of pertinent information from every chosen study. This encompasses, though is not exclusively confined to, the study's aims, predictive process monitoring, and explainability approaches, results, and issues or contextual factors that are essential for comprehending its impact on the discipline. In order to uphold uniformity and precision, a standardized data extraction form is employed, encompassing all essential particulars that will subsequently prove pivotal in the synthesis and analysis stages.

After the completion of data collection, the research proceeds to the subsequent phase, known as a qualitative synthesis of studies. In this phase, the primary methodology employed is template analysis proposed by King (2012), which offers a flexible yet methodical framework for the thematic arrangement and understanding of textual data \cite{king2012doing}. The process of template analysis encompasses a series of fundamental stages, beginning with formulating an initial template. This template serves as a fundamental structure for systematically classifying and arranging the collected data. The initial template has been formulated based on a comprehensive analysis of the review objectives and a preliminary examination of the predictive process monitoring and XAI methods described in the Background section. This approach ensures that the starting point is firmly rooted in the established research body while allowing for potential adjustments and refinements.

 \begin{figure}[t!]
 \centering
 \includegraphics[width=0.7\textwidth]{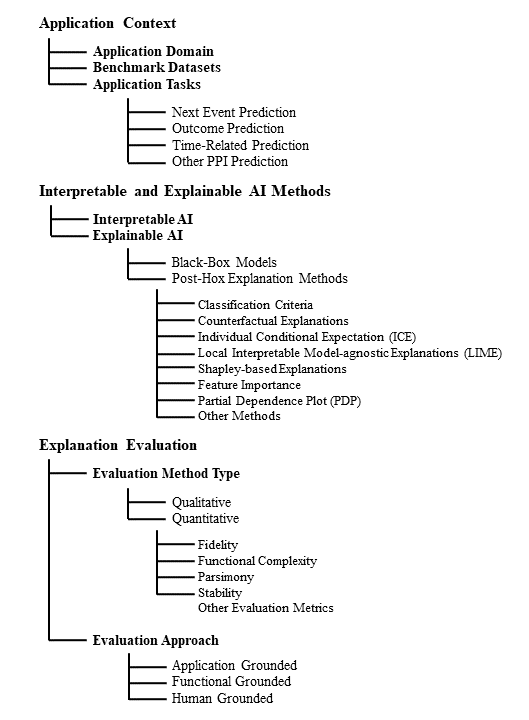}
 \caption{Template for the analysis approach of retrieved literature.}
 \label{figure_analysis_template}
 \end{figure}

The template undergoes iterative revisions and refinements as we progressively explore the data. The process entails encoding the collected data derived from the conducted studies into a designated template, alongside the discernment and identification of novel themes or sub-themes that manifest throughout the analysis. The emergence of these novel perspectives necessitates the adaptation of the framework, be it through incorporating additional themes, refining preexisting ones, or reconfiguring the overall structure to more accurately capture the emerging connections and patterns. The aforementioned iterative process persists until a state of stability is attained in the template, wherein it effectively encapsulates the various themes and patterns that emerge from the produced data.

The final template subsequently functions as a foundational framework for the comprehensive combination of the data (see Figure \ref{figure_analysis_template}). In this analysis, we engage in the interpretation and discourse surrounding the various themes present while concurrently establishing connections among relevant studies. Our aim is to identify patterns, discrepancies, and emerging trends within the body of literature. The synthesis presented herein not only elucidates the present state of scholarly inquiry but also imparts a nuanced comprehension of the trajectory, obstacles, and prospective avenues for advancement within the field.

\subsection{Study Selection}\label{selection}
The selection process commenced with the identification of records through an extensive search across multiple databases and registers, including ACM, AIS, IEEE, Science Direct, Springer Link, and additional backward and forward searches. This initial step yielded a total of 1,071 records. Each record was subjected to a careful screening process. Titles and abstracts were reviewed to determine their relevance to the study's inclusion criteria, which led to the exclusion of 980 records for reasons not meeting the specified research scope and objectives. Consequently, 91 reports were selected for retrieval and further evaluation. In the eligibility assessment phase, the full texts of these 91 reports were meticulously examined to ascertain their suitability for inclusion in the review. During this phase, reports were excluded based on predefined exclusion criteria, labeled as EX1 through EX4, which represented various rationales for ineligibility, such as irrelevance to the research questions, methodological shortcomings, or lack of empirical data. This resulted in the exclusion of an additional 24 reports. The culmination of this rigorous selection process was the inclusion of 67 studies in the final review. These studies were deemed to align closely with the research objectives and met all the criteria set forth for the systematic review. No additional reports of included studies were identified, affirming the thoroughness of the search and selection strategy.

\begin{figure}[b]
 \centering
 \includegraphics[width=0.8\textwidth]{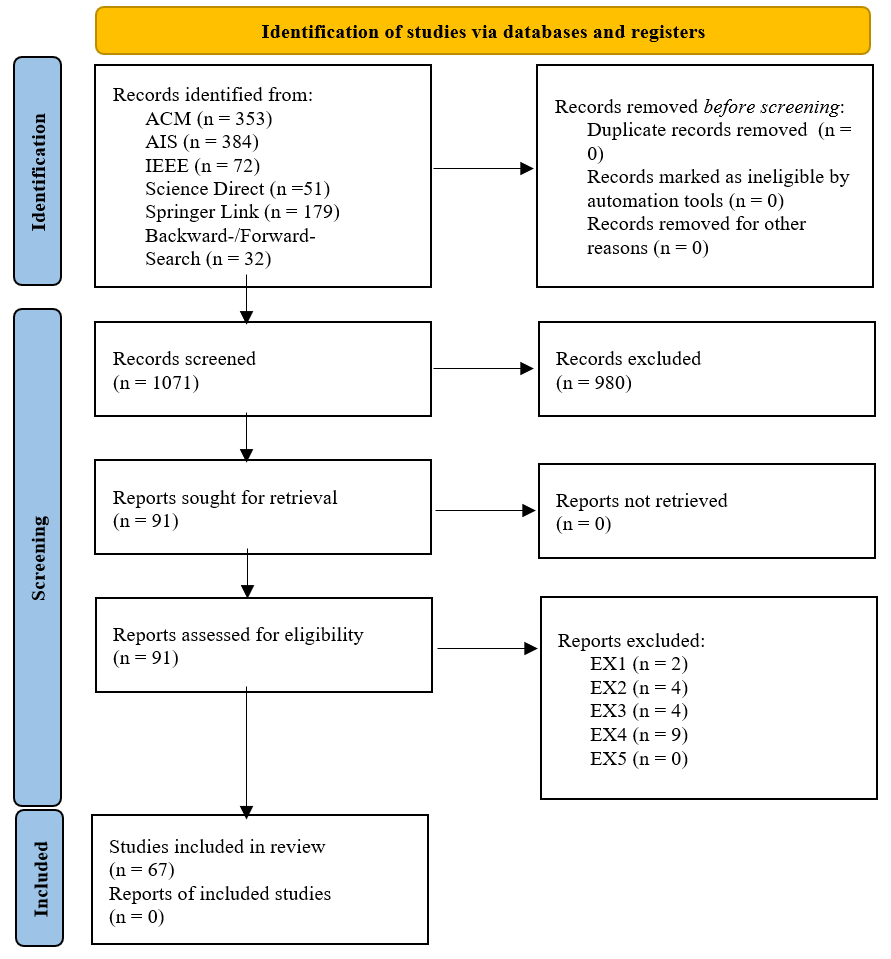}
 \caption{Flowchart depicting the retrieval and selection of retrieved publications, following the PRISMA approach.}
 \label{flow_search_figure}
 \end{figure}

The transparent and systematic approach to the study selection, as evidenced by the PRISMA flow diagram (see Figure \ref{flow_search_figure}), ensures a high level of confidence in the comprehensiveness and relevance of the studies included in this review. This process underscores the robustness and reliability of the findings and discussions that will be presented, providing a solid foundation for the synthesis and analysis that follow.

\hfill

\section{Results}\label{sec4}
This section presents the findings of the literature review and is systematically divided into four key subsections, each addressing a specific aspect of our research.
Section \ref{descriptive_analysis} delves into the analysis of metadata derived from our research data. It presents the patterns and trends that emerged from examining the metadata, offering insights into the characteristics and distribution of the data utilized in our study. Section \ref{application_context} explores the application domains of the approaches described in the found articles. This part provides an in-depth look at the implications of our results in different domains and highlights prevalent application fields. Section \ref{interpretable_explainable_AI_Methods} analyzes the employed approaches and ML models as well as the utilized explanation methods. Lastly, Section \ref{explanation_evaluation} examines the evaluation of employed explanation techniques. Each of these subsections collectively contributes to a comprehensive understanding of our research findings, offering a multi-faceted view of our study's impact and significance.

\subsection{Descriptive Analysis}\label{descriptive_analysis}
\begin{figure}[b]
 \centering
 \includegraphics[width=0.70\textwidth]{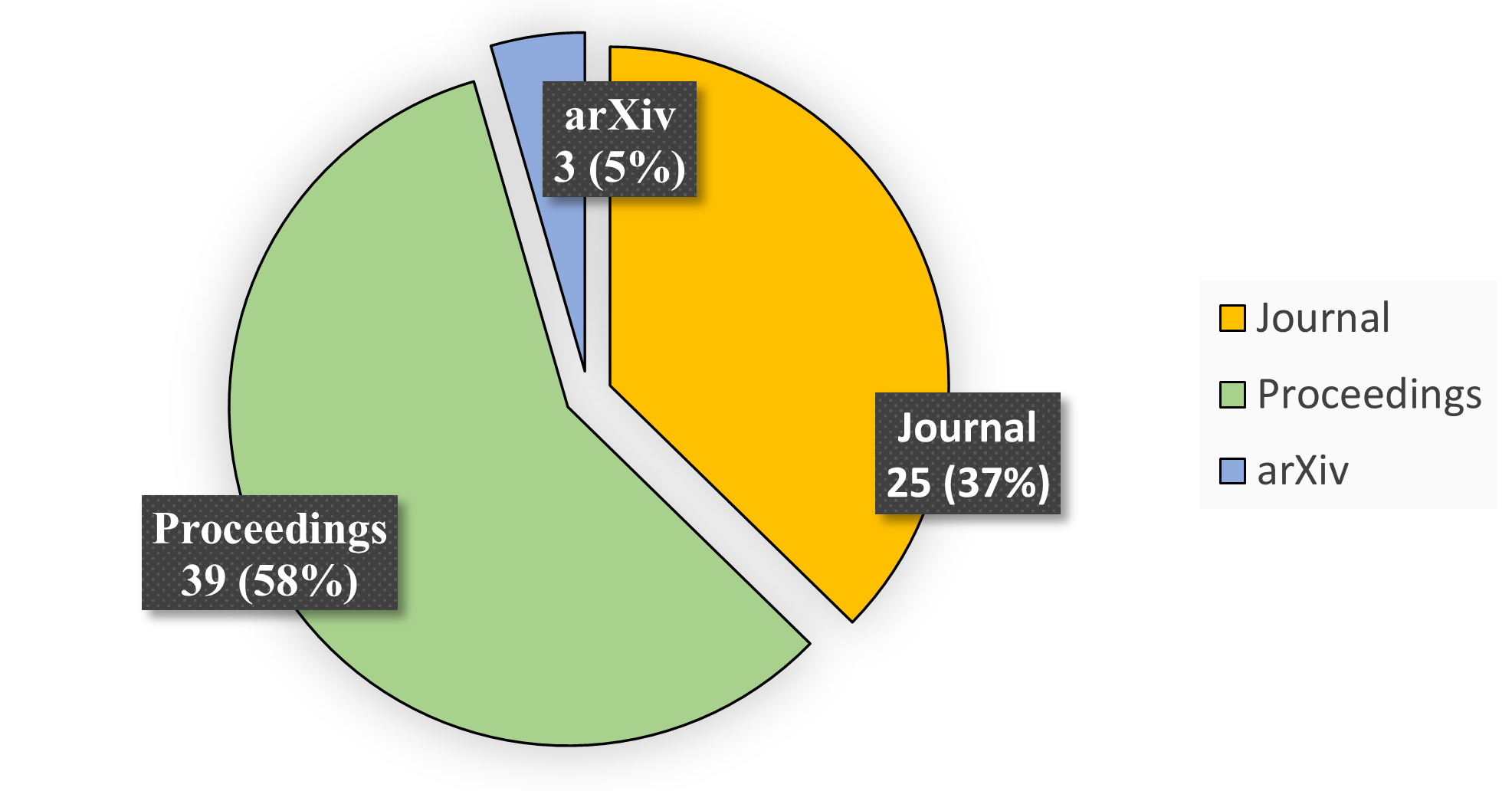}
 \caption{Number of identified publications per publication outlet.}
 \label{publications_per_medium_figure}
 \end{figure}

For the analysis of metadata, the publication outlet and year as well as corresponding keywords were examined: Regarding the publication outlet, 39 out of the 67 articles were published in conference proceedings, 25 in journals and three via arXiv, as visualized via pie-chart in Figure \ref{publications_per_medium_figure}. The analyzed publications media, with the exception of arXiv, are known to be peer-reviewed sources, as per SLR standards. However, as a result of the backwards-search, articles published via arXiv were included as well for the purpose of completeness.

\begin{figure}[t]
 \centering
 \includegraphics[width=0.95\textwidth]{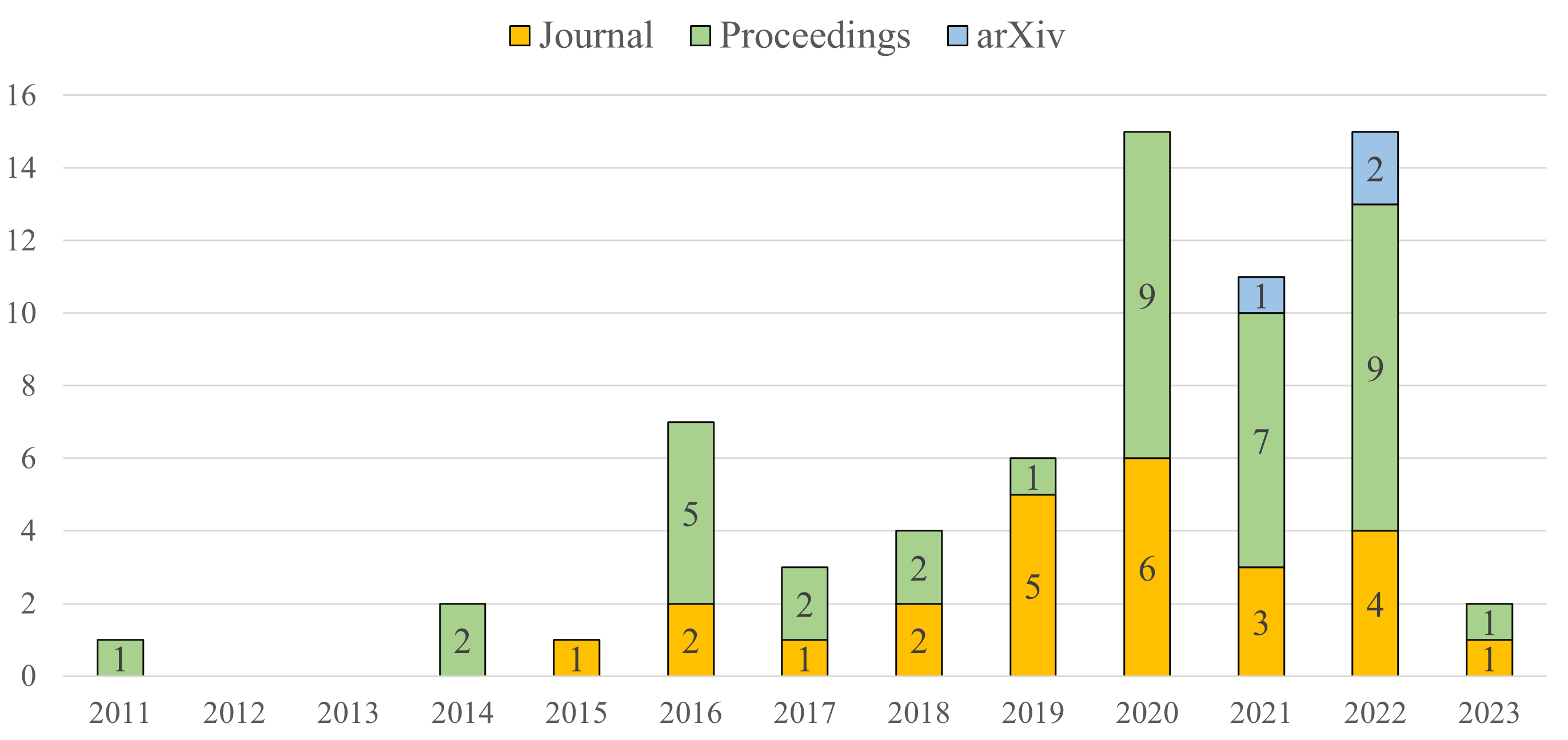}
 \caption{Number of identified publications per publication outlet grouped by year of publication.}
 \label{publications_per_year_figure}
 \end{figure}

Regarding the publishing date of identified literature, Figure \ref{publications_per_year_figure} depicts the publications per year and publication medium in the form of a stacked bar chart. On closer examination, a spike in the amount of publications around the year 2020 can be observed. The majority of the literature was published in 2020 and onward (43 out of 67 articles), with 2020 and 2022 being the years with the most publications (15 out of 67 articles), suggesting an upward trend in the adoption of interpretable ML approaches for predictive process monitoring. 

\begin{figure}[h]
 \centering
 \includegraphics[width=11cm]{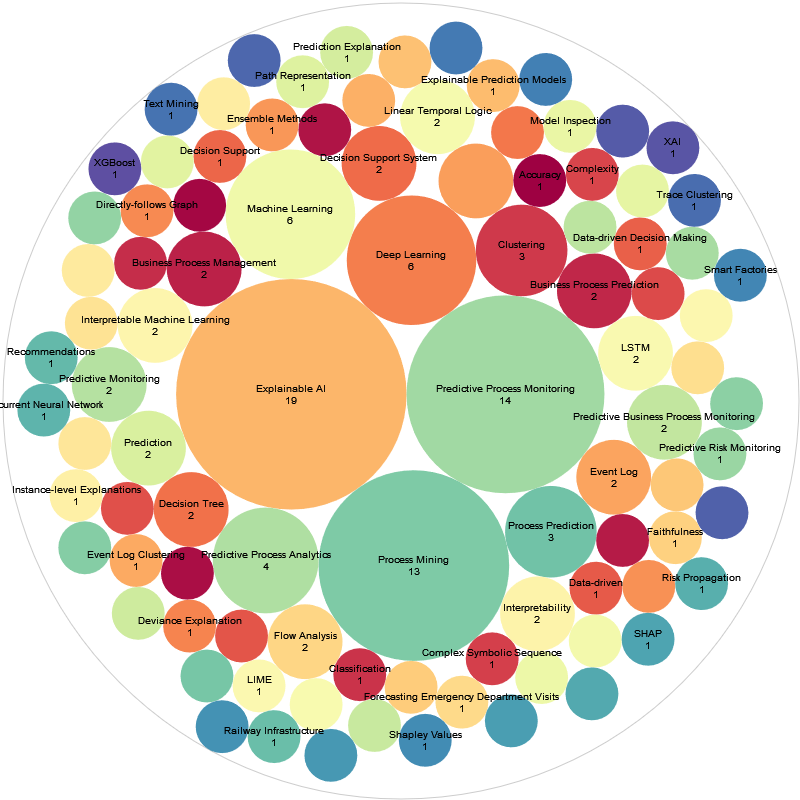}
 \caption{Circle packing diagram of usage and frequency of article keywords.}
 \label{keywords_circle_packing_figure}
 \end{figure}

For the analysis of keywords, either chosen by the authors or proposed by the publication outlet, the identified articles were visualized via a circle packing chart depicted in Figure \ref{keywords_circle_packing_figure}, illustrating the keywords and corresponding frequency of occurrence. Visually, larger circles depict a more frequent use of the keyword (or phrase) within the circle compared to smaller circles, e.g. "Predictive process monitoring" occurred in 14 publications. It is noteworthy that different representations of the same concepts were used, such as "Explainable Artificial Intelligence" and "Explainable AI" being used as a key-phrase to depict the domain of an article. For the visualization, keywords describing the same concepts were grouped together under a single keyword. The analysis of keywords shows, that approximately half of the articles (31 out of 67) aimed to contribute directly to the XAI domain. Considering the search process for relevant literature, the variety in employed keywords and their formulation outlines the challenges in the adequate formulation of search queries in order to cover various iterations of the terminology specific to the XAI-domain.\\

\subsection{Application Context}\label{application_context}
This subsection delineates the examination of the retrieved publications, encompassing a descriptive analysis, identified application domains, utilized benchmark datasets and central application tasks. First, the descriptive analysis examines the distribution of publications across various publication outlets and the prevalence of specific keywords, followed by the analysis of the application domains. Next, datasets employed for the conception and evaluation of the proposed methodologies are examined. Lastly, an analysis on the underlying application tasks is being performed, presenting the most prevalent types among the retrieved articles. For the remainder of this section, we refer to the Tables \ref{table_application_domain_A} and \ref{table_application_domain_B} for a detailed documentation of application domains and tasks as well as utilized datasets identified in the retrieved literature. The following subsections offer a comprehensive and coherent overview of the current research landscape in XAI, emphasizing its relevance and applicability in the field of ML and process analytics.

\subsubsection{Application Domain}\label{application_domain}
For the identification of the application domain, the properties of the utilized datasets as well as explicit statements by the authors were analyzed and aggregated. These characteristics allow for the distinction between domain-agnostic and domain-specific applications of the presented approaches, and give insight into the work areas covered in the literature.

As the most prevalent application domains finance (represented in 40 out of 67 articles), healthcare (18 out of 67 articles), customer support services (18 out of 67 articles) and manufacturing (9 out of 67 articles) were identified. Approximately half of the publications (30 out of 67) were assessed as domain-agnostic, due to their independence towards the field of application, thus, demonstrating the transferability of the underlying methodology. 
Considering the close relationship between application domain and the datasets utilized for model training and evaluation, the following section provides an deeper analysis of the benchmark datasets leveraged in the retrieved articles.\\

\subsubsection{Benchmark Datasets}\label{benchmark_datasets}
\begin{figure}[h]
 \centering
 \includegraphics[width=12cm]{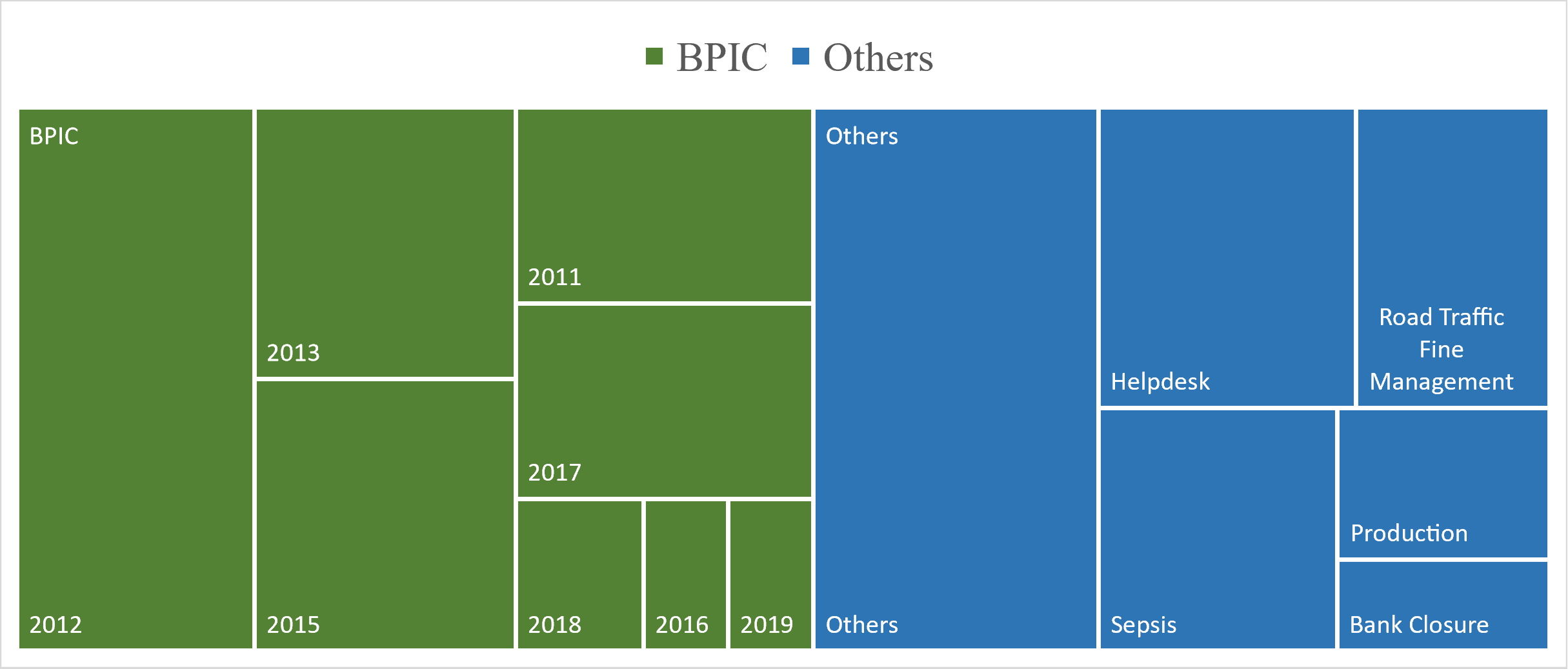}
 \caption{Treemap diagram representing the usage of various event logs.}
 \label{event_logs_analysis_figure}
 \end{figure}

Since the employed datasets dictate the possible application domains, examining the utilized event logs not only provides information about the presented application domains, but also about the degree of transferability and adaptiveness of the approaches presented in the analyzed articles. Figure \ref{event_logs_analysis_figure} is a treemap diagram depicting the usage of various event logs, arranged by the frequency in ascending order, with the size of each area correlating to the amount of publications that used the corresponding dataset. 

The event logs were separated into two groups: One group encompasses the BPIC event logs, the other includes the rest of the datasets (datasets that have not been used by two or more publications were allocated to the "Others" category). The BPIC 2011 event log is taken from an Academic Hospital and is therefore allocated to the healthcare-domain, BPIC 2012 and 2017 cover loan-application processes and were allocated to the finance sector, similar to BPIC 2016 which deals with employee insurance, BPIC 2018 which deals with financing applications and BPIC 2019 which pertains to the processing of invoices. Although the BPIC 2013 dataset originally stems from an automobile company, the event log itself is restricted to incident management, and is therefore allocated to customer support services. BPIC 2015 deals with building permit applications and was not allocated a dedicated domain due to its low representation in the retrieved articles.
The rest of the datasets and their corresponding application domains were categorized as follows: Bank Account Closure was allocated to finance, Helpdesk to customer support services, Production to manufacturing, and Sepsis to healthcare. The Road Traffic Fine Management falls under law enforcement, but was not allocated to a dedicated application domain due to low representation. Datasets that have not been explicitly described in this section were either of synthetic nature or inaccessible, in which case the authors' statements about the application domain were incorporated for this categorization.

In the found literature, the Business Process Intelligence Challenges (BPIC) dataset catalogue is being employed predominantly, with 45 out of 67 articles (~67\% of found publications) using at least one of the provided datasets. The usage of the same data over various publications facilitates the benchmarking of results, which is one of the main reasons for the utilization of the BPIC event logs stated within the articles. Another reason is the open-source nature of these datasets, making them easily accessible to the public and therefore contributing to the transparency and replicability of the presented approaches. Lastly, all of the BPIC datasets are real-life event logs, facilitating approaches that aim to be grounded in reality. Regarding the frequency of utilization, the BPIC 2012 event log was employed the most (utilized in 20 out of 67 articles), thus contributing to the finance domain being the prevalent application domain. Almost half of the publications (30 out of 67) implemented their approach on at least two event logs from differing application domains, demonstrating the domain-agnostic nature of the underlying appraoch. Regardless of domain, 32 out of 67 articles evaluated their approaches on two or more datasets, examining the robustness of the proposed methodology across data from different sources.

\subsubsection{Application Tasks}\label{application_tasks}
The utilization of certain ML models depends heavily on the prediction tasks at hand. Especially in process prediction, there are prevalent prediction tasks that entail certain types of explanations as well as corresponding explanation objects and subjects.
Since the prediction task is integral for the selection of the employed ML model, and therefore on the objectives of explanation methods, this section presents the prediction tasks of the retrieved articles and categorizes prediction tasks into the following four groups: The first group encompasses the prediction of the next event of an unfinished process trace. This is the case for non-deterministic processes where the expression of certain features, context factors as well as events within the unfinished trace itself influence what activity is going to be executed next. The second group deals with the prediction of process outcomes, for example the prediction of anomalies within a process at runtime or the allocation of events or traces to predefined categories. The third and fourth group deals with the prediction of process performance indicators, with the third group particularly encompassing predictions of time-related PPI, such as the remaining time until completion for an event or an unfinished process trace. The fourth group is comprised of PPI prediction tasks unrelated to time, such as the prediction of context variables, costs and others.

First, publications that aimed for the prediction of the next event are being presented, followed by those that predicted process outcomes. Afterwards, articles that predicted time-related process performance indicators are being presented, and lastly, literature with other process performance indicators prediction tasks.\\

\noindent\textbf{Next Event Prediction}\label{next_event_prediction}\\
The prediction of the upcoming events, given an unfinished process trace, is the second most prevalent application task within the retrieved literature (22 our of 67 articles) and is mostly found in the context of optimizing the production process by being able to plan ahead. Articles that aimed at solving this problem type were Agarwal et al. \cite{agarwal2022process}, Böhmer \& Rinderle-Ma \cite{bohmer2018probability}, Böhmer \& Rinderle-Ma \cite{Bohmer2020}, Brunk et al. \cite{Brunk2021}, De Leoni et al. \cite{DeLeoni2015}, Gerlach et al. \cite{gerlach2022inferring}, Hanga et al. \cite{Hanga2020}, Hsieh et al. \cite{Hsieh2021}, Lakshmanan et al. \cite{Lakshmanan2011}, Maggi et al. \cite{Maggi2014}, Mayer, Mehdiyev \& Fettke \cite{mayer2021manufacturing}, Rehse et al. \cite{Rehse2019}, Savickas \& Vasilecas \cite{savickas2018belief}, Sindghatta et al. \cite{Sindhgatta2020a}, Tama et al.\cite{tama2020empirical}, Unuvar et al. \cite{Unuvar2016}, Verenich et al. \cite{Verenich2017}, Verenich et al. \cite{Verenich2019b}, Weinzierl et al. \cite{Weinzierl2020}, Wickramanayake et al. \cite{Wickramanayake2022}, Wickramanayake et al. \cite{Wickramanayake2022a} and Zilker et al. \cite{zilker2023best}. Among these articles, next event prediction is often accompanied by other prediction tasks, with Lakshmanan et al. \cite{Lakshmanan2011} and Unuvar et al. \cite{Unuvar2016} being examples that aim not only at predicting the next but also the following activities up until the end of a given trace. Maggi et al. \cite{Maggi2014} describe next event prediction as a byproduct of their approach, although not the primary goal of their work, similar to Verenich et al. (2017)\cite{Verenich2017} and Verenich et al. (2019) \cite{Verenich2019b} where next event prediction is realized as an implicit task by allocating probabilities to reachable states of a given process trace.\\

\noindent\textbf{Process Outcome Prediction}\label{process_outcome_prediction}\\
Although the details of process outcome prediction depend heavily on the application context at hand, this application task is most prevalent among the analyzed literature, with 41 articles out of 67 confronting outcome prediction tasks: Agarwal et al. \cite{agarwal2022process}, Böhmer \& Rinderle-Ma \cite{bohmer2020mining}, Bukhsh et al. \cite{AllahBukhsh2019}, Conforti et al. \cite{Conforti2016}, De Koninck et al. \cite{DeKoninck2017}, De Leoni et al. \cite{DeLeoni2015}, De Oliveira et al. \cite{de2020optimization}, De Oliveira et al. \cite{de2020automatic}, Di Francescomarino et al. \cite{DiFrancescomarino2016}, Di Francescomarino et al. \cite{Francescomarino2019}, Folino et al. \cite{Folino2017},  Galanti et al. \cite{Galanti2020},  Galanti et al. \cite{Galanti2022}, Garcia-Banuelos et al. \cite{garcia2017complete}, Harl et al. \cite{Harl2020}, Horita et al. \cite{horita2016goal}, Huang et al. \cite{Huang2022}, Irarrazaval et al. \cite{irarrazaval2021telecom}, Khemiri \& Pinaton \cite{khemiri2018improving}, Lakshmanan et al. \cite{Lakshmanan2011},  Maggi et al. \cite{Maggi2014}, Mehdiyev \& Fettke \cite{Mehdiyev2021}, Mehdiyev \& Fettke \cite{Mehdiyev2020a}, Mehdiyev \& Fettke \cite{Mehdiyev2020}, Mehdiyev et al. \cite{mehdiyev2021explainable}, Ouyang et al. \cite{Ouyang2021}, Pasquadibisceglie et al. \cite{Pasquadibisceglie2021}, Pauwels \& Calders \cite{pauwels2019detecting}, Pauwels \& Calders \cite{pauwels2019anomaly}, Prasisdis et al. \cite{prasidis2021handling}, Rehse et al. \cite{Rehse2019}, Rizzi et al. \cite{Rizzi2020}, Savickas \& Vasilecas \cite{savickas2014business}, Sindghatta et al. \cite{Sindhgatta2020}, Stevens \& de Smedt \cite{Stevens2022}, Stevens et al. \cite{stevens2022assessing}, Stevens et al. \cite{Stevens2022a}, Teinemaa et al. \cite{Teinemaa2016}, Velmurugan et al. \cite{Velmurugan2021a}, Velmurugan et al. \cite{Velmurugan2021} and Verenich et al. \cite{Verenich2016}. 

Among the variety problems addressed by the authors, trace classification or clustering as addressed by De Koninck et al. \cite{DeKoninck2017}, De Oliveira et al. \cite{de2020optimization}, De Oliveira et al. \cite{de2020automatic}, Di Francescomarino et al. \cite{DiFrancescomarino2016}, Di Francescomarino et al. \cite{Francescomarino2019}, Folino et al. \cite{Folino2017} and Verenich et al. \cite{Verenich2016} as well as anomaly detection as addressed by Böhmer \& Rinderle-Ma \cite{bohmer2020mining}, Garcia-Banuelos et al. \cite{garcia2017complete}, Irarrazaval et al. \cite{irarrazaval2021telecom}, Pauwels \& Calders \cite{pauwels2019detecting}, Pauwels \& Calders \cite{pauwels2019anomaly} are documented as prevalent prediction tasks. Other tasks encompass maintenance prediction (Bukhsh et al. \cite{AllahBukhsh2019}), risk detection (Conforti et al. \cite{Conforti2016}), insurance reclamation (De Leoni et al. \cite{DeLeoni2015}).\\

\noindent\textbf{Time Related Prediction}\label{time_related_prediction}\\
The prediction of indicators for process performance is a regression task in need of a process expert in order to define and/or identify impactful variables in order to yield results relevant to the underlying production process. The prediction of time-related process parameters, such as the remaining time for the processing of a given event or trace or the prediction of execution times for certain activities, are the most prevalent tasks within the analyzed literature and are being addressed by the following articles:
Böhmer \& Rinderle-Ma \cite{bohmer2018probability}, Böhmer \& Rinderle-Ma \cite{Bohmer2020}, Cao et al. \cite{cao2023business}, Cao et al. \cite{cao2023explainable}, De Leoni et al. \cite{DeLeoni2015}, Galanti et al. \cite{Galanti2020},  Galanti et al. \cite{Galanti2022}, Mayer, Mehdiyev \& Fettke \cite{mayer2021manufacturing}, Ouyang et al. \cite{Ouyang2021}, Padella et al. \cite{padella2022explainable}, Polato et al. \cite{polato2018time}, Rehse et al. \cite{Rehse2019}, Sindghatta et al. \cite{Sindhgatta2020}, Toh et al. \cite{toh2022improving}, Verenich et al. \cite{Verenich2017} and Verenich et al. \cite{Verenich2019b}.
Exemplary time-related prediction problems encompass the prediction of the timestamp of the next event (Böhmer \& Rinderle-Ma \cite{Bohmer2020}), the  prediction of execution times of activities for a given trace (Rehse et al. \cite{Rehse2019}, Verenich et al. (2017) \cite{Verenich2017} and Verenich et al. (2019) \cite{Verenich2019b}) and the prediction of remaining time until completion for a given unfinished trace (De Leoni et al. \cite{DeLeoni2015}, Ouyang et al. \cite{Ouyang2021} and Sindghatta et al. \cite{Sindhgatta2020}), with the works of Galanti et al. (2020) \cite{Galanti2020} and Galanti et al. (2022) \cite{Galanti2022} also predicting the total cost of the relevant trace.\\

\begin{table}[p]
\begin{center} 
\begin{minipage}{330pt}
\setlength{\tabcolsep}{2pt}
\caption{Categorization of application task, application domain and utilized event log in the found literature}
\label{table_application_domain_A}
\begin{tabular}{c|cc|cc|ccccc|c|cccccccc|ccccc|c}
\toprule
& \multicolumn{4}{c|}{\footnotesize Application}                         & \multicolumn{6}{c|}{\footnotesize Application}   & \multicolumn{14}{c}{\footnotesize Event Log}\\
& \multicolumn{4}{c|}{\footnotesize Task}                                & \multicolumn{6}{c|}{\footnotesize Domain}        & \multicolumn{14}{c}{\footnotesize}\\

& \multicolumn{2}{c|}{\footnotesize Class.} & \multicolumn{2}{c|}{\footnotesize Regr.} & \multicolumn{5}{c|}{} &          & \multicolumn{8}{c|}{\footnotesize BPIC} & \multicolumn{5}{c|}{\footnotesize Misc.} &\\

Publication &
\rot{\footnotesize Next Event} & \rot{\footnotesize Process Outcome} & \rot{\footnotesize Time-related} & \rot{\footnotesize Other PPI} & 
\rot{\footnotesize Customer Support} & \rot{\footnotesize Finance} & \rot{\footnotesize Healthcare} & \rot{\footnotesize Manufacturing} & \rot{\footnotesize Others} & \rot{\footnotesize Domain Agnostic} &            
\rot{\footnotesize BPIC 2011} & \rot{\footnotesize BPIC 2012} & \rot{\footnotesize BPIC 2013} & \rot{\footnotesize BPIC 2015} & \rot{\footnotesize BPIC 2016} & \rot{\footnotesize BPIC 2017} & \rot{\footnotesize BPIC 2018} & \rot{\footnotesize BPIC 2019} &       
\rot{\footnotesize Bank Account Closure} & \rot{\footnotesize Helpdesk} &\rot{\footnotesize Production} & \rot{\footnotesize Road Traffic Fine Man.} & \rot{\footnotesize Sepsis}  & \rot{\footnotesize Other}\\
\botrule
\end{tabular}
\end{minipage}
\end{center}
\end{table}

\begin{table}[p]
\begin{center} 
\begin{minipage}{330pt}
\setlength{\tabcolsep}{2pt}
\caption{Categorization of application task, application domain and utilized event log in the found literature}
\label{table_application_domain_B}
\begin{tabular}{c|cc|cc|ccccc|c|cccccccc|ccccc|c}
\toprule
& \multicolumn{4}{c|}{\footnotesize Application}                         & \multicolumn{6}{c|}{\footnotesize Application}   & \multicolumn{14}{c}{\footnotesize Event Log}\\
& \multicolumn{4}{c|}{\footnotesize Task}                                & \multicolumn{6}{c|}{\footnotesize Domain}        & \multicolumn{14}{c}{\footnotesize}\\

& \multicolumn{2}{c|}{\footnotesize Class.} & \multicolumn{2}{c|}{\footnotesize Regr.} & \multicolumn{5}{c|}{} &          & \multicolumn{8}{c|}{\footnotesize BPIC} & \multicolumn{5}{c|}{\footnotesize Misc.} &\\

Publication &
\rot{\footnotesize Next Event} & \rot{\footnotesize Process Outcome} & \rot{\footnotesize Time-related} & \rot{\footnotesize Other PPI} & 
\rot{\footnotesize Customer Support} & \rot{\footnotesize Finance} & \rot{\footnotesize Healthcare} & \rot{\footnotesize Manufacturing} & \rot{\footnotesize Others} & \rot{\footnotesize Domain Agnostic} &            
\rot{\footnotesize BPIC 2011} & \rot{\footnotesize BPIC 2012} & \rot{\footnotesize BPIC 2013} & \rot{\footnotesize BPIC 2015} & \rot{\footnotesize BPIC 2016} & \rot{\footnotesize BPIC 2017} & \rot{\footnotesize BPIC 2018} & \rot{\footnotesize BPIC 2019} &       
\rot{\footnotesize Bank Account Closure} & \rot{\footnotesize Helpdesk} &\rot{\footnotesize Production} & \rot{\footnotesize Road Traffic Fine Man.} & \rot{\footnotesize Sepsis}  & \rot{\footnotesize Other}\\
\botrule
\end{tabular}
\end{minipage}
\end{center}
\end{table}

\noindent\textbf{Other Process Performance Indicator Predictions}\label{other_ppi_prediction}\\
Apart from the prediction of time-related PPI, a variety of other PPI-related prediction tasks has been documented for the work of Bayomie et al. \cite{bayomie2022improving}, Coma-Puig \&  Carmona \cite{coma2022non}, Fu et al. \cite{fu2021modeling}, Galanti et al. (2020) \cite{Galanti2020}, Galanti et al. (2022) \cite{Galanti2022}, Mayer, Mehdiyev \& Fettke \cite{mayer2021manufacturing} and Petsis et al. \cite{Petsis2022}. In particular, Bayomie et al. \cite{bayomie2022improving} define and predict a numeric indicator for event-case correlation, Coma-Puig \&  Carmona \cite{coma2022non} quantify and predict non-technical energy loss, while Fu et al. \cite{fu2021modeling} does the same for customer experience.
Apart from remaining time, Galanti et al. (2020) \cite{Galanti2020} and Galanti et al. (2022) \cite{Galanti2022} also predict the costs associated with the process, similar to Mayer, Mehdiyev \& Fettke \cite{mayer2021manufacturing}. Along with Petsis et al. \cite{Petsis2022} predicting the number of patient visits, it is observable that the prediction of other process performance indicators pertain to relevant application tasks for the respective application domain.\\

 \begin{figure}[h]
 \centering
 \includegraphics[width=\textwidth]{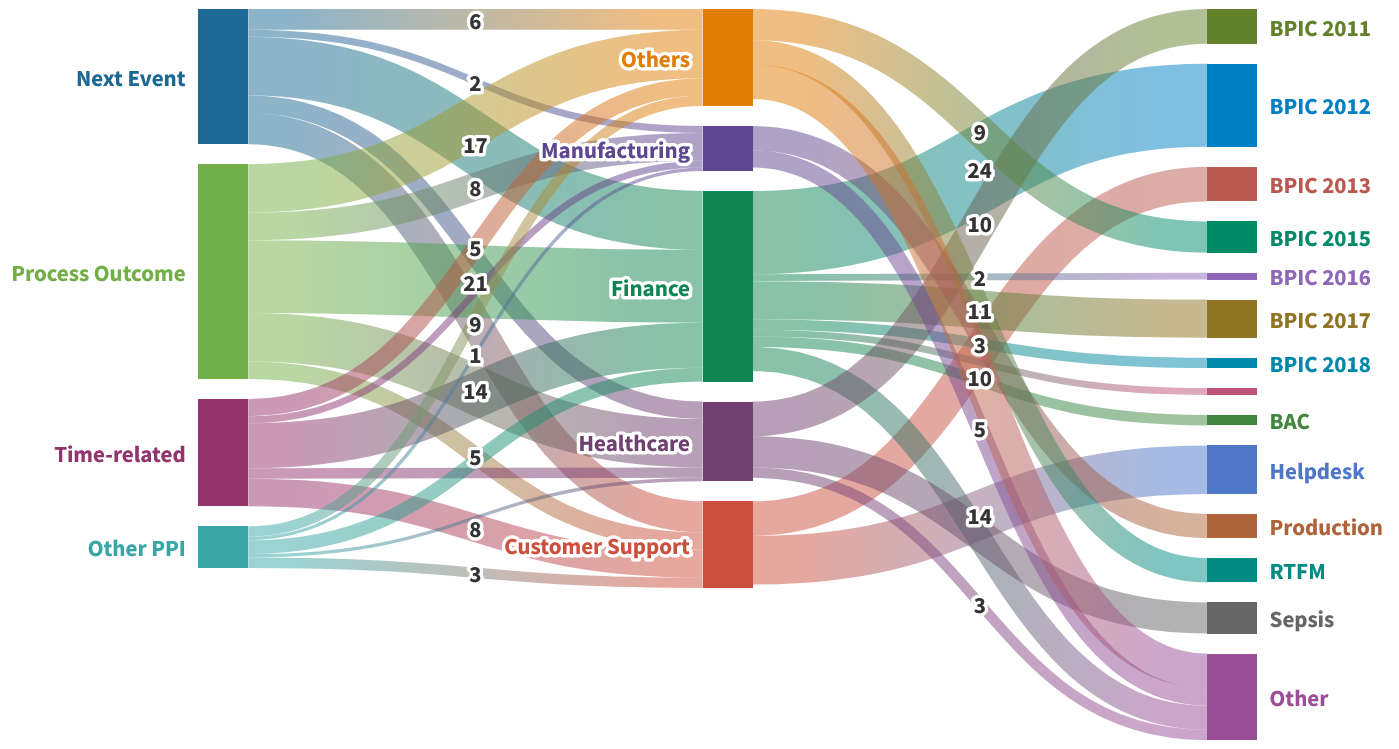}
 \caption{Sankey-Diagram representing the application task, the application domains and the corresponding application datasets. The line width represents the amount of scenarios found in the analyzed literature.}
 \label{tasks_and_domain_figure}
 \end{figure}

It is evident that classification tasks were prevalent in the found literature, with 22 articles addressing next event prediction, 41 articles covering process outcome prediction, totaling at 58 articles. Within regression tasks (20 out of 67 articles), predicting the time-related PPI for a given event or trace was aimed for in 16 articles, with 8 articles predicting other process related PPI. For a more comprehensive analysis, the Sankey diagram in Figure \ref{tasks_and_domain_figure} illustrates the relationship between the application tasks, application domain and employed datasets of the analyzed articles. This figure demonstrates the prevalence of process outcome tasks, followed by next event and time-related predictions. The finance domain is being addressed the most, which can be traced back to its predominant representation in the BPIC datasets, with the BPIC 2012 event log being utilized in approximately one third of retrieved articles (24 out of 67).

\subsection{Interpretable and Explainable AI Methods} \label{interpretable_explainable_AI_Methods}
This section will categorize the found publications based on the characteristics of the employed AI methods in the context of XAI. First, a general classification of prevalent ML models is being presented, delineating the differentiation between interpretable AI and explainable AI, followed by the analysis of the approaches employed by the authors of found publications with regards to the utilized models and explanation methods.

The in-depth literature review talks about and sorts common models in the ML field by how easy they are to understand, especially when it comes to algorithmic transparency, decomposability, and simulatability \cite{BARREDOARRIETA202082}. Bayesian models or networks, decision trees, general additive models, k-nearest neighbors, linear regression and logistic regression models, as well as rule-based learners, were characterized as providing an acceptable level of functional transparency by design and, thus, not necessarily needing post-hoc explanations. This is due to the fact that simulatability is realistically possible by humans, although many models require some level of decomposition in order to be analyzed or need additional mathematical tools in order to comprehend model behavior in the context of algorithmic transparency. All of the above models are, therefore, characterized as interpretable models. In contrast, deep learning (DL) models (like convolutional neural networks (CNN), deep feedforward neural networks (DFNN) or recurrent neural networks (RNN)), gradient boosting models (GBM), support vector machines (SVM) and ensemble approaches do not provide any inherent algorithmic transparency, decomposability or simulatability within reasonable human means. Therefore, the quality of explainability of these models is directly dependent upon the employed post-hoc explanation method. Therefore, these models lack fundamental transparency and are classified here as explainable models. Tables \ref{table_model_expl_A} and \ref{table_model_expl_B} provide a categorization of the given articles for this literature review, based on the employed ML method and characteristics of the provided explanations, in particular explanation scope, relation towards the corresponding model and output format, and is used as an orientation for the remainder of this section.\\

\subsubsection{Interpretable AI} \label{interpretable_AI}
The following retrieved publications integrated and evaluated interpretable AI as a means to solve PPM-related tasks: 
Agarwal et al. \cite{agarwal2022process}, Bayomie et al. \cite{bayomie2022improving}, Böhmer \& Rinderle-Ma \cite{bohmer2018probability}, Böhmer \& Rinderle-Ma \cite{bohmer2020mining}, Böhmer \& Rinderle-Ma \cite{Bohmer2020}, Brunk et al. \cite{Brunk2021}, Bukhsh et al. \cite{AllahBukhsh2019}, Conforti et al. \cite{Conforti2016}, De Leoni et al. \cite{DeLeoni2015}, De Oliveira et al. \cite{de2020optimization}, Di Francescomarino et al. \cite{DiFrancescomarino2016}, Di Francescomarino et al. \cite{Francescomarino2019}, Folino et al. \cite{Folino2017}, Fu et al. \cite{fu2021modeling}, Garcia-Banuelos et al. \cite{garcia2017complete}, Horita et al. \cite{horita2016goal}, Irarrazaval et al. \cite{irarrazaval2021telecom}, Khemiri \& Pinaton \cite{khemiri2018improving}, Lakshmanan et al. \cite{Lakshmanan2011}, Maggi et al. \cite{Maggi2014}, Mayer, Mehdiyev \& Fettke \cite{mayer2021manufacturing}, Pauwels \& Calders \cite{pauwels2019detecting}, Pauwels \& Calders \cite{pauwels2019anomaly}, Polato et al. \cite{polato2018time}, Prasisdis et al. \cite{prasidis2021handling}, Savickas \& Vasilecas \cite{savickas2014business}, Savickas \& Vasilecas \cite{savickas2018belief}, Stevens \& de Smedt \cite{Stevens2022}, Stevens et al. \cite{stevens2022assessing}, Stevens et al. \cite{Stevens2022a}, Tama et al.\cite{tama2020empirical} and Unuvar et al. \cite{Unuvar2016}. Among above articles, decision trees were the most prevalent approaches with representation in 13 articles, followed by bayesian networks employed in six articles and linear or logistic regression approaches represented in 5 articles. Among the remaining 16 interpretable methods, clustering approaches like k-means or heuristic rule-based clustering were leveraged, as well as methods that combine several interpretable AI methodologies. 

Regarding decision trees, Bukhsh et al. \cite{AllahBukhsh2019}, De Leoni et al. \cite{DeLeoni2015}, Di Francescomarino et al. (2016) \cite{DiFrancescomarino2016}, Di Francescomarino et al. (2019) \cite{Francescomarino2019}, Horita et al. \cite{horita2016goal}, Irarrazaval et al. \cite{irarrazaval2021telecom}, Khemiri \& Pinaton \cite{khemiri2018improving}, Lakshmanan et al. \cite{Lakshmanan2011}, Maggi et al. \cite{Maggi2014}, Stevens \& de Smedt \cite{Stevens2022}, Stevens et al. \cite{Stevens2022a} and Unuvar et al. \cite{Unuvar2016} leveraged this interpretable approach in the context of a variety of PPM tasks. In particular, Bukhsh et al. \cite{AllahBukhsh2019} compared three different ML methods with a decision tree implementing CART (Classification and Regression Trees) in the fashion of Breiman et al. \cite{Breiman1983ClassificationAR} being one of those methods. The aim was predicting maintenance of railway switches within the domain of railway infrastructure. The model has been implemented with two other methods (random forest and gradient boosting trees) and evaluated based on each model's measured accuracy, F-1, kappa, and misclassification scores. De Leoni et al. \cite{DeLeoni2015} implemented the proposed framework as a plug-in for the ProM Framework (van Dongen et al. \cite{10.1007/11494744_25}) and, given an event log as input, mines a process model yielding either a corresponding Decision Tree (C4.5, see Quinlan \cite{C4.5} and Mitchell \cite{Mitchell1997}) or Regression Tree (RepTree, see Witten \cite{Witten2011}). The authors advise splitting the event log that is given as input into use-case-specific clusters to increase homogeneity within the process behavior within the mined models, increasing the validity of the resulting models. As application tasks, the presented framework allows for predicting upcoming events, process outcomes or the remaining time until process completion and was evaluated on the BPIC 2016 event log. Di Francescomarino et al. (2016) \cite{DiFrancescomarino2016} presented a predictive process monitoring framework, that has been implemented in the ProM framework as an Operational Support provider in order to be able to perform during runtime. The proposed framework encodes a given event log either frequency- or sequence-based, passes it to a clustering method (either Agglomorative Clustering, DBSCAN or K-Means Clustering), and eventually employs either Decision Trees or random forests as classification models. The framework allows for manual optimization of certain hyperparameters and the final models are being evaluated based on their accuracy, earliness (refering to how early within a given trace a prediction can be formed), failure rate as well as computation time. This approach has been evaluated on the BPIC 2011 and BPIC 2015 event logs with the aim to predict certain process outcomes. Di Francescomarino et al. (2019) \cite{Francescomarino2019} presented another predictive process monitoring framework, similar to their work in 2016, also implemented in the ProM framework as an Operational Support provider in order to be able to perform during runtime. The framework distinguishes itself from the previously presented article by proposing two clustering methods: model-based clustering as proposed by Fraley \& Raftery  \cite{model_based_clustering} for frequency-based encoding of the event log and DBSCAN for sequence-based encoding. Lakshmanan et al. \cite{Lakshmanan2011} presented a binary decision tree implementation using C4.5 on a synthetically generated event log and evaluated the model by its accuracy, simulating an insurance claim scenario. Given a trace from an unfinished process the model is expected to predict the process outcome. Maggi et al. \cite{Maggi2014} presented a framework that classifies traces of a given event log based on the application scenario and use case, and then proceeds to build a corresponding C4.5 decision tree in order to predict the next event or process outcome for new traces. This approach has been implemented in the ProM framework as an operational support provider in order to be able to perform during runtime and has been evaluated on the BPIC 2011 event log. The resulting models have been evaluated on the accuracy, AUROC, F-1-scores, false positive (FPR) and true positive rates (TPR), positive predictive values (PPV), and Receiving Operating Characteristics (ROC).

Bayesian networks, as implemented in Brunk et al. \cite{Brunk2021}, Pauwels \& Calders \cite{pauwels2019detecting}, Pauwels \& Calders \cite{pauwels2019anomaly}, Prasisdis et al. \cite{prasidis2021handling}, Savickas \& Vasilecas \cite{savickas2014business} and Savickas \& Vasilecas \cite{savickas2018belief}, were leveraged for a transparent approach towards event log analysis, confronting tasks like next event or process outcome prediction and anomaly detection. As exemplary work, Brunk et al. \cite{Brunk2021} employed a Dynamic Bayesian Network with a manually defined structure in order to predict the next event within a given trace of an event log. This approach aimed at differentiating attributes of the event log that are the cause or the effect of a given process and was evaluated on the BPIC 2012 and BPIC 2013 data sets. For benchmarking, implementations of probabilistic finite automata and n-grams were utilized to compare accuracy and various approaches presented in other publications for the given event logs. 

Linear or logistic regression approaches were leveraged by Agarwal et al. \cite{agarwal2022process}, Bukhsh et al. \cite{AllahBukhsh2019}, De Leoni et al. \cite{DeLeoni2015}, Stevens \& de Smedt \cite{Stevens2022}, Stevens et al. \cite{stevens2022assessing} and Teinemaa et al. \cite{Teinemaa2016}. Agarwal et al. \cite{agarwal2022process} proposed a decision support system employing logistic regression for process outcome and next event prediction, while Stevens \& de Smedt \cite{Stevens2022}, Stevens et al. \cite{stevens2022assessing} presented a methodology for process outcome prediction with a strong focus on the evaluation of model explanations. Teinemaa et al. \cite{Teinemaa2016} presented an approach of predicting the process outcome for two real-life event logs from the domain of finance (dept recovery and lead-to-contract processes) by employing techniques from text-mining in order to encode process traces. A logistic regression model has been utilized as a classifier for said task and was evaluated on their computation time, F-1- and earliness scores. However, the authors did not include specific results for the proposed approach, justified by it being outperformed by the random forest model on any employed evaluation metric.

The remainder of articles employing interpretable approaches, particularly Bayomie et al. \cite{bayomie2022improving}, Böhmer \& Rinderle-Ma \cite{bohmer2018probability}, Böhmer \& Rinderle-Ma \cite{bohmer2020mining}, Böhmer \& Rinderle-Ma \cite{Bohmer2020}, Conforti et al. \cite{Conforti2016}, De Oliveira et al. \cite{de2020optimization}, De Oliveira et al. \cite{de2020automatic}, Folino et al. \cite{Folino2017}, Fu et al. \cite{fu2021modeling}, Garcia-Banuelos et al. \cite{garcia2017complete}, Horita et al. \cite{horita2016goal}, Irarrazaval et al. \cite{irarrazaval2021telecom}, Maggi et al. \cite{Maggi2014}, Mayer, Mehdiyev \& Fettke \cite{mayer2021manufacturing}, Polato et al. \cite{polato2018time}, Stevens \& de Smedt \cite{Stevens2022}, Stevens et al. \cite{Stevens2022a} and Tama et al. \cite{tama2020empirical}, cover a variety of (mixed) approaches in order to tackle a multitude of PPM prediction tasks. As an example, Böhmer \& Rinderle-Ma \cite{Bohmer2020} introduced sequential prediction rules in the context of next event prediction and evaluated their approach ("LoGo") on the BPIC 2012 and Helpdesk data sets based on the mean absolute error and accuracy, comparing their approach to LSTM and RNN models. These rules are applied to specific traces of a given event log, aiming to predict the next activity on a general level for said trace. If no general rules exist for said trace, then probability based heuristics are employed as a classifier, comparing the given trace to similar traces from historic data. Conforti et al. \cite{Conforti2016} present "PRISM", an approach aiming at detecting risks in real-time during process execution by using dedicated sensors. Conceptionally, a process model is being developed for the use case incorporating risk-annotations. Sensors are designed on top of this model, process predefined risk conditions and trigger an alarm to the process administrator if certain conditions are met. The approach also incorporates a similarity measure between instances and, thus, any instance that has been identified as containing a risk will lead to similar instances being identified as well before the corresponding sensors are able to conduct further analysis. Folino et al. \cite{Folino2017} present a rule based clustering approach employing propositional patterns. This approach was evaluated on the BPIC 2013 event log and compared to an implementation of M5Rules (see Holmes et al. \cite{M5Rules}) on interestingness and explanation complexity.\\

\subsubsection{Explainable AI} \label{explainable_AI}
With opaque models being predominantly utilized in the found literature (40 out of 67 articles) compared to interpretable models (32 out of 67), unveiling their inner working necessitates an explicit post-hoc explanation approach. This section provides an overview of the predominant black-box model types used in the retrieved literature as well as the explanation methods utilized.\\

\noindent\textbf{Black-Box Models}\\
This section covers black-box approaches found in the analyzed literature. First, articles employing deep learning methods are presented, followed by gradient boosting models, random forests and, lastly, models that fall in neither of these prevalent categories. Publications already covered in section \ref{interpretable_AI} employing black-box models are briefly mentioned where appropriate.

The following publications employed deep learning models: Cao et al. \cite{cao2023business}, Cao et al. \cite{cao2023explainable}, Galanti et al. \cite{Galanti2020}, Galanti et al. \cite{Galanti2022}, Gerlach et al. \cite{gerlach2022inferring}, Hanga et al. \cite{Hanga2020}, Harl et al. \cite{Harl2020}, Hsieh et al. \cite{Hsieh2021}, Huang et al. \cite{Huang2022}, Mayer, Mehdiyev \& Fettke \cite{mayer2021manufacturing}, Mehdiyev \& Fettke \cite{Mehdiyev2021}, Mehdiyev \& Fettke \cite{Mehdiyev2020a}, Mehdiyev \& Fettke \cite{Mehdiyev2020}, Mehdiyev et al. \cite{mehdiyev2021explainable}, Pasquadibisceglie et al. \cite{Pasquadibisceglie2021}, Rehse et al. \cite{Rehse2019}, Sindghatta et al. \cite{Sindhgatta2020a}, Stevens \& de Smedt \cite{Stevens2022}, Stevens et al. \cite{stevens2022assessing}, Stevens et al. \cite{Stevens2022a}, Weinzierl et al. \cite{Weinzierl2020}, Wickramanayake et al. \cite{Wickramanayake2022}, Wickramanayake et al. \cite{Wickramanayake2022a}, and Zilker et al. \cite{zilker2023best}, encompassing deep neural networks (DNN), recurrent neural networks (RNN), long short-term memory (LSTM) RNN as well as combined approaches. As exemplary work, Mehdiyev \& Fettke \cite{Mehdiyev2021}, \cite{Mehdiyev2020a}, and \cite{Mehdiyev2020} employed DNN in all three of their publications, focusing on performant models and post-hoc explainability. Galanti et al. (2020) \cite{Galanti2020} utilized an LSTM, while Hanga et al. \cite{Hanga2020} performed a comparative analysis between a conventional and a bidirectional LSTM, and compared both against the results of similar studies. Hsieh et al. \cite{Hsieh2021} proposed an approach that leverages an ensemble of a DNN and an LSTM, introducing "DiCE4EL" - a modified implementation of "DiCE" (see Mothilal et al. \cite{DiCE}), applicable to event logs. Huang et al. \cite{Huang2022} utilized for their "LORELEY" approach an LSTM to be applicable to event logs. Rehse et al. \cite{Rehse2019} utilize an LSTM, exploring potentials of explainability within process prediction in the context of the DFKI-Smart-Lego-Factory (see Rehse et al. \cite{DFKI_Smart_Lego_Factory}). Sindghatta et al. (2020b) \cite{Sindhgatta2020a} present an approach utilizing a Bidirectional LSTM in one case and an ensemble of two Bidirectional LSTM in two other cases, depending on the application task. Weinzierl et al. \cite{Weinzierl2020} presented "XNAP", a model-specific approach that employs a Bidirectional LSTM RNN that is able to propagate feature relevance scores from one layer to another. Wickramanayake et al. (2022a) \cite{Wickramanayake2022} build upon the approach from Sindghatta et al. (2020b) \cite{Sindhgatta2020a}, presenting two architectures, both of which use ensembles of bidirectional LSTM models in similar fashion. Wickramanayake et al. (2022b) \cite{Wickramanayake2022a} proposed an explanation framework in the context of the Wickramanayake et al. (2022a) \cite{Wickramanayake2022} publication, employing the previously mentioned model architecture. Stevens et al. \cite{Stevens2022a} and Stevens \& de Smedt \cite{Stevens2022} utilized LSTM models as well, the former in conjunction with an XGBoost model for benchmarking, the latter in conjunction with a CNN and random forest models in order to perform a qualitative and quantitative comparison of their approach for a variety of models.

The following articles leveraged gradient boosting models in their methodology, either as the central model of the proposed approach or for comparative analysis against other models: Bukhsh et al. \cite{AllahBukhsh2019}, Coma-Puig \& Carmona \cite{coma2022non},  Galanti et al. \cite{Galanti2022}, Mayer, Mehdiyev \& Fettke \cite{mayer2021manufacturing}, Mehdiyev et al. \cite{mehdiyev2021explainable}, Ouyang et al. \cite{Ouyang2021}, Padella et al. \cite{padella2022explainable}, Petsis et al. \cite{Petsis2022}, Sindghatta et al. \cite{Sindhgatta2020}, Stevens \& de Smedt \cite{Stevens2022}, Stevens et al. \cite{Stevens2022a}, Toh et al. \cite{toh2022improving}, Velmurugan et al. \cite{Velmurugan2021a}, Velmurugan et al. \cite{Velmurugan2021} and Verenich et al. \cite{Verenich2019b}. In particular, Stevens \& de Smedt \cite{Stevens2022} implemented a generalized logistic rule model (GLRM), a logistic regression model and a logit leaf model as white-box models with a CNN, a LSTM, a random forest as well as an XGBoost model as black-box models, and evaluated their approach on the BPIC 2011, BPIC 2015, Production and Sepsis event logs. The employed models aimed at predicting process outcomes and their predictive performance were evaluated based on their Area under the Receiving Operating Characteristics Curve (AUROC). This approach has been implemented in the context of a guideline ("X-MOP") proposed by the authors, aiming at selecting the appropriate model for the corresponding application task and scenario. Similarly, Stevens et al. \cite{Stevens2022a} employ a GLRM in the context of comparing white-box and black-box approaches based on their functional complexity, monotonicity and parsimony. Velmurugan et al. \cite{Velmurugan2021} aimed at evaluating the stability of LIME and SHAP explanation methods in the context of process outcome prediction. The approach employed logistic regression as the white-box model and compared it to an XGBoost black-box model, evaluated on the BPIC 2012, Production and Sepsis event logs with, taking various data encoding methods into account. Ouyang et al. \cite{Ouyang2021} and Petsis et al. \cite{Petsis2022} and Sindghatta et al. (2020a) \cite{Sindhgatta2020}, Velmurugan et al. \cite{Velmurugan2021a} and Verenich et al. (2019) \cite{Verenich2019b} employed XGBoost models for the evaluation of post-hoc explainability techniques.
 
Regarding the use of random forest in the found literature, the following publications leveraged this model type predominantly for process outcome prediction tasks, either by itself or in comparison with other ML-methods: Bukhsh et al. \cite{AllahBukhsh2019}, Rizzi et al. \cite{Rizzi2020}, Stevens \& de Smedt \cite{Stevens2022},  Stevens et al. \cite{stevens2022assessing}, Teinemaa et al. \cite{Teinemaa2016}, Verenich et al. \cite{Verenich2016}, Verenich et al. \cite{Verenich2017}. In particular, Bukhsh et al. \cite{AllahBukhsh2019} utilized a random forest approch, apart from the decision tree and graident boosting trees, comparing the performance of all three models. Rizzi et al. \cite{Rizzi2020} propose an approach employing random forest and retraining the model based on the analysis of explanations provided for the former model. Teinemaa et al. \cite{Teinemaa2016} implemented their proposed approach with a random forest model as an alternative, comparing its performance against logistic regression. Verenich et al. \cite{Verenich2016} presented an approach that builds a random forest on top of an event log after the corresponding traces have been clustered using one of two proposed clustering algorithms. Similarly, Verenich et al. \cite{Verenich2017} employ a random forest as a classifier on the level of each activity within a trace after allocating said trace to a discovered process model based on the given event log. 

The following publications employed models that do not fall in any category of the previously presented ones: De Koninck et al. \cite{DeKoninck2017} proposed an approach in the context of trace clustering, employing a modified "Search for Explanations for Clusters of Process Instances" (SECPI) (De Weerdt \& vanden Broucke \cite{SECPI}) architecture, utilizing a Support Vector Machines for each identified cluster to find the minimal set of features that allow a given instance to stay in its allocated cluster. Verenich et al. \cite{Verenich2016}, Verenich et al. \cite{Verenich2017} and Verenich et al. \cite{Verenich2019b}, respectively added a clustering and two process model discovery components to their approach, thus adding an interpretable layer on top of their black-box approaches.\\

\noindent\textbf{Post-Hoc Explanation Methods}
Post-hoc explanation methods exhibit a variety of differences, depending on the model that is explained, as well as the application context and PPM task that is being tackled. In particular, the following characteristics are differentiated: Regarding explanation scope, local and global explanations are distinguished, with the former focusing on explanations pertaining to individual model predictions and the latter referring to the general workings of the examined model. The model relation differentiates between model-specific explanation methods, which leverage the intricacies of the model methodology, and model-agnostic explanation methods, which can be applied regardless of the utilized model. Lastly, the output format of the explanation can be in numeric, textual, rule-based, or visual form as well as a mixture thereof.

Local XAI methods focus on revealing the relevance of variables for predictions on a single data point. In contrast to global explanations, local explanations do not necessarily uncover general model behavior but provide valuable insight into specific prediction instances. Nevertheless, depending on the underlying data and use case, local predictions for similar instances have the capability to capture model behavior within a given locality. Hence, these methods allow for an extrapolation of local findings in order to derive insights about global model behaviour, with some local methods laying the foundation for global explanation methods.\\

\begin{figure}[h]
 \centering
 \includegraphics[width=10cm]{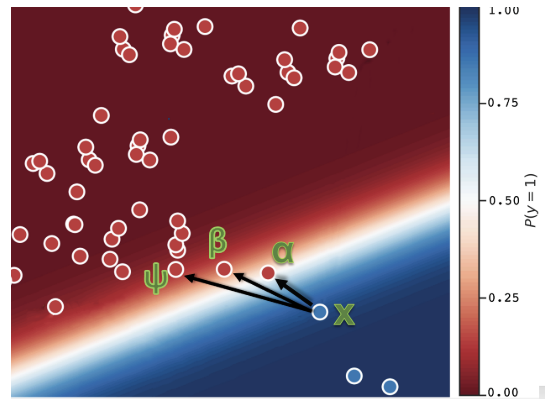}
 \caption{General example for a Counterfactual Explanation as it is implemented by  Hsieh et al. \cite{Hsieh2021}, demonstrating the original instance as well as the found counterfactual instances that flip the outcome label.}
 \label{Counterfactual_explanations_general_example}
 \end{figure}

\begin{figure}[h]
\centering
\begin{subfigure}[b]{0.5\textwidth}
   \includegraphics[width=1\linewidth]{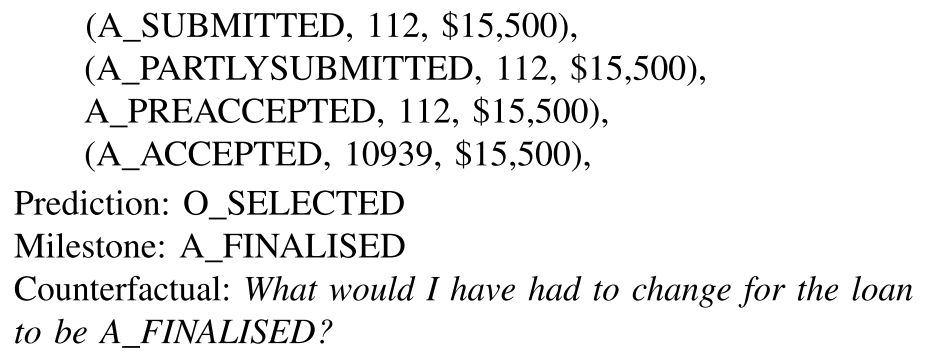}
   \caption{}
\end{subfigure}

\begin{subfigure}[b]{1\textwidth}
   \includegraphics[width=1\linewidth]{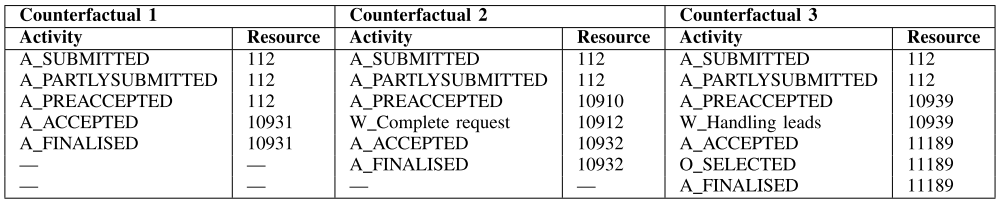}
   \caption{}
\end{subfigure}

\caption[LIME Examples]{
Example for a Counterfactual Explanation as it is implemented by Hsieh et al. \cite{Hsieh2021}.
(a) demonstrates the original instance, whereas
(b) demonstrates the counterfactual explanations and the features that have been altered to achieve the desired prediction - in this case, the acceptance of a loan of \$15,500.
}
\label{Counterfactual_explanations_example}
\end{figure}

\noindent\textbf{Counterfactual Explanations}\\
The Counterfactual explanation is a contrastive method of providing insight by presenting conditions, specifically certain variable values, under which the prediction score would exceed or fall below a certain threshold compared to its original score. These explanations aim to identify the least amount of intervention in order to flip a prediction label for classification tasks or bring the prediction score across a certain threshold for regression tasks. Counterfactual explanations have informative characteristics and provide, depending on the ability to manipulate certain variables, actionable advice for attaining specific prediction scores. However, the fact that an exhaustive search for counterfactual explanations is likely to suffer from a combinatorial explosion for categorical variables and that it can be expected to find various such explanations necessitates an implementation that is suitable for its corresponding application context. Figure \ref{Counterfactual_explanations_general_example} is an example of a visual counterfactual explanation from Hsieh et al. \cite{Hsieh2021}, illustrating the original instance as well as counterfactual instances with modified feature values that result in the prediction score exceeding a given threshold. In a similar fashion, Hsieh et al. \cite{Hsieh2021} implemented counterfactual explanations using a tabular visualization for the altered features of the counterfactual explanations, as seen in Figure \ref{Counterfactual_explanations_example}. Similar approaches towards counterfactual explanations can be found in De Koninck et al. \cite{DeKoninck2017}, Huang et al. \cite{Huang2022}, Mayer, Mehdiyev \& Fettke \cite{mayer2021manufacturing} and Padella et al. \cite{padella2022explainable}.\\

\begin{figure}[h]
 \centering
 \includegraphics[width=\textwidth]{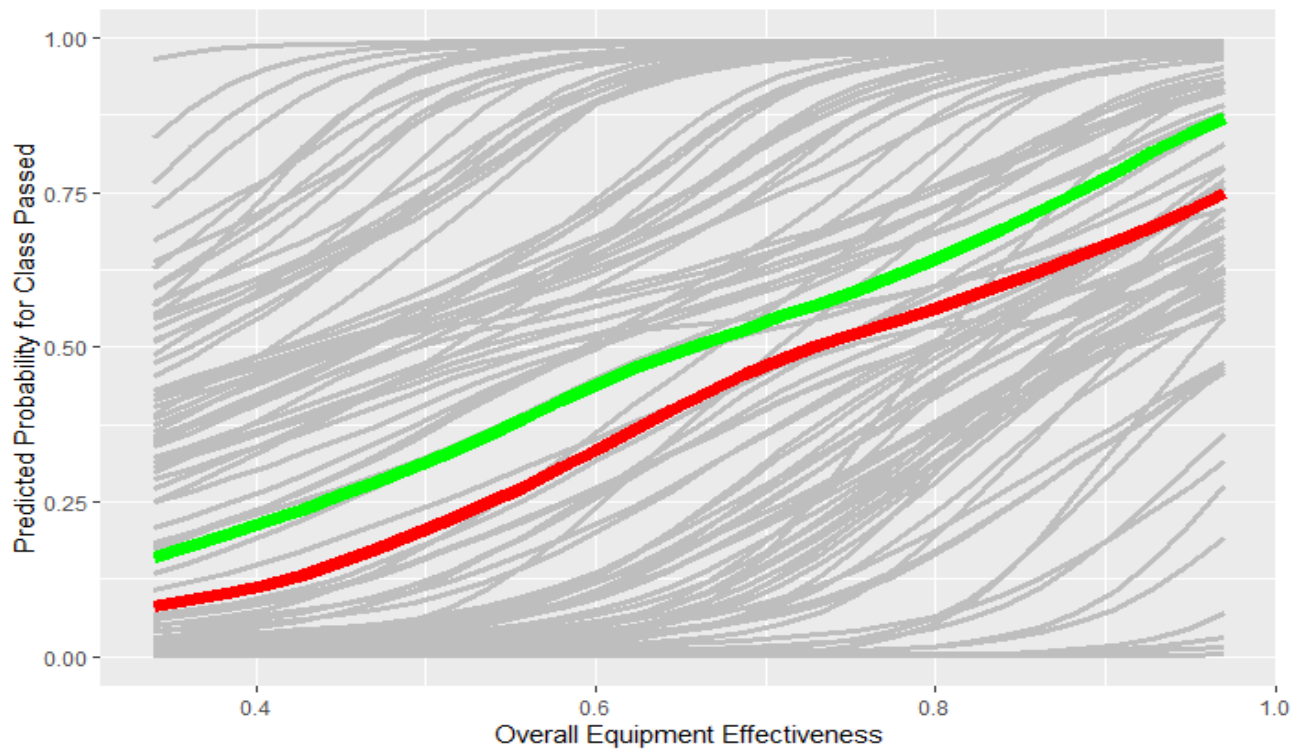}
 \caption{Example of an Individual Conditional Expectation plot as it was implemented in Mehdiyev \& Fettke \cite{Mehdiyev2020}, with the green line depicting a true positive instance and the red line depicting a true negative instance. }
 \label{ice_example_figure}
 \end{figure}

\noindent\textbf{Individual Conditional Expectation (ICE)}\\
Individual Conditional Expectation (ICE) plots are a model-agnostic approach and conceptually similar to PDP in that they illustrate the impact of an iterated feature for a single data point, whereas PDP present the mean response for said feature over all data points. Algorithmically, the value of a given variable of an instance is being iterated over its observed values for categorical variables or over certain ranges for numerical variables, and the resulting change in the prediction score is being captured. In practice, ICE plots can be visualized for an individual instance or for a group of instances in a single plot, depending on the use case, although the latter approach qualifies as a global explanation. Figure \ref{ice_example_figure} is an example of an ICE plot from Mehdiyev \& Fettke (2020c) \cite{Mehdiyev2020}, illustrating the changes of prediction scores for each single instances within a group (visualized as one line per instance) across value changes of the "Overall Equipment Effectiveness" variable. One of the advantages of ICE plots over PDP is that a visualization such as Figure \ref{ice_example_figure} facilitates the identification of and differentiation between global and local model behaviour. Other publications employing ICE are Mayer, Mehdiyev \& Fettke \cite{mayer2021manufacturing} and Mehdiyev et al. \cite{mehdiyev2021explainable}.\\

\begin{figure}[h]
\centering
\begin{subfigure}[b]{0.75\textwidth}
   \includegraphics[width=1\linewidth]{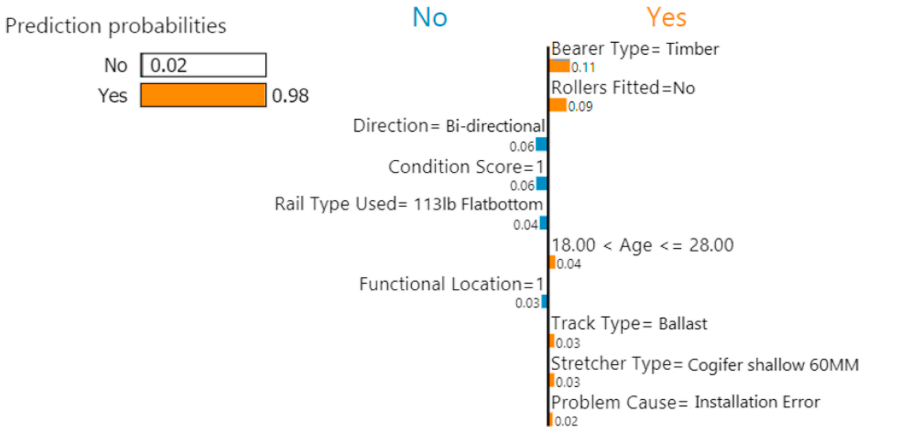}
   \caption{}
   \label{fig:Ng1} 
\end{subfigure}

\begin{subfigure}[b]{0.75\textwidth}
   \includegraphics[width=1\linewidth]{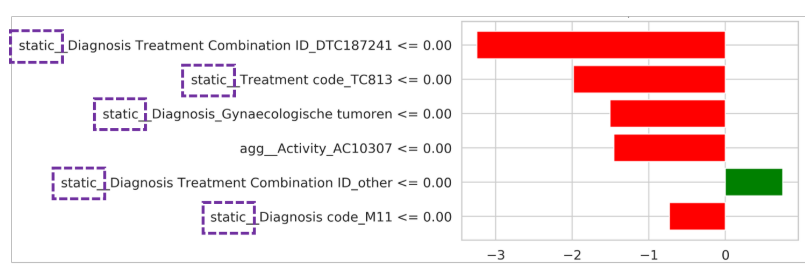}
   \caption{}
   \label{fig:Ng2}
\end{subfigure}

\caption[LIME Examples]{
Example for LIME as it is implemented by (a) Bukhsh et al. \cite{AllahBukhsh2019} and (b) Ouyang et al. \cite{Ouyang2021}, illustrating feature with positive impact on the prediction score on the right-hand side using (a) orange/ (b) green bars and features with negative impact on the prediction score on the left-hand side using (a) blue/ (b) red bars. The length of the colored bars represents the impact of the feature on the prediction score, with the corresponding numerical value as labels in (a) or visibile on the x-Axis in (b).
}
\label{LIME_examples_figure}
\end{figure}

\noindent\textbf{Local Interpretable Model-agnostic Explanations (LIME)}\\
LIME as per Ribeiro et al. \cite{LIME_Ribeiro} rely on a specific implementation of surrogate models that specialize on mimicking the behaviour of an underlying model for a certain locality within the data set. For this approach, sufficiently explainable surrogate models are being trained on a data set with iterated feature values and corresponding prediction scores to these modified instances, provided by the underlying model for which the surrogate is being built, in order to learn model behaviour within a certain locality. Provided that the surrogate model attains a sufficiently high local fidelity, the inherent explainability of the surrogate model allows for explanations for the behaviour of the underlying model in said locality by proxy, i.e the impact of features on the prediction score. 
In the analyzed literature, the following articles employed LIME as an explanation technique: Bukhsh et al. \cite{AllahBukhsh2019} (see Figure \ref{LIME_examples_figure} (a)), Mayer, Mehdiyev \& Fettke \cite{mayer2021manufacturing}, Mehdiyev et al. \cite{mehdiyev2021explainable}, Ouyang et al. \cite{Ouyang2021} (Figure \ref{LIME_examples_figure} (b)), Rizzi et al. \cite{Rizzi2020}, Sindghatta et al. \cite{Sindhgatta2020}, Velmurugan et al. \cite{Velmurugan2021a} and Velmurugan et al. (2021a) \cite{Velmurugan2021}. In particular, Velmurugan et al. (2021b) \cite{Velmurugan2021} employed LIME in the style of Visani et al. \cite{stable_LIME_Visani}, measuring the feature contribution via LIME over ten surrogate models in order to capture the stability of LIME explanations. Although LIME can leverage the advantages of interpretable models, the identification and clustering of instances that would fall into a specific locality is a significant obstacle for non-image data and depends heavily on the underlying use case. Mehdiyev \& Fettke \cite{Mehdiyev2021} implemented a modified, model-specific approach, conceptually based on LIME and K-LIME (Hall et al. \cite{k_lime_Hall}) using neural codes from the last hidden layer of a DNN as a vector for distance calculation between instances, thus identifying localities based on the learned instance representations of the underlying model. Rehse et al. \cite{Rehse2019} mention a similar approach, leveraging the neural codes from the last hidden layer of a DNN in order to identify localities for specific instances, however, the authors do not specify the interpretable model that was used for their method.\\

\begin{figure}[h]
 \centering
 \includegraphics[width=11cm]{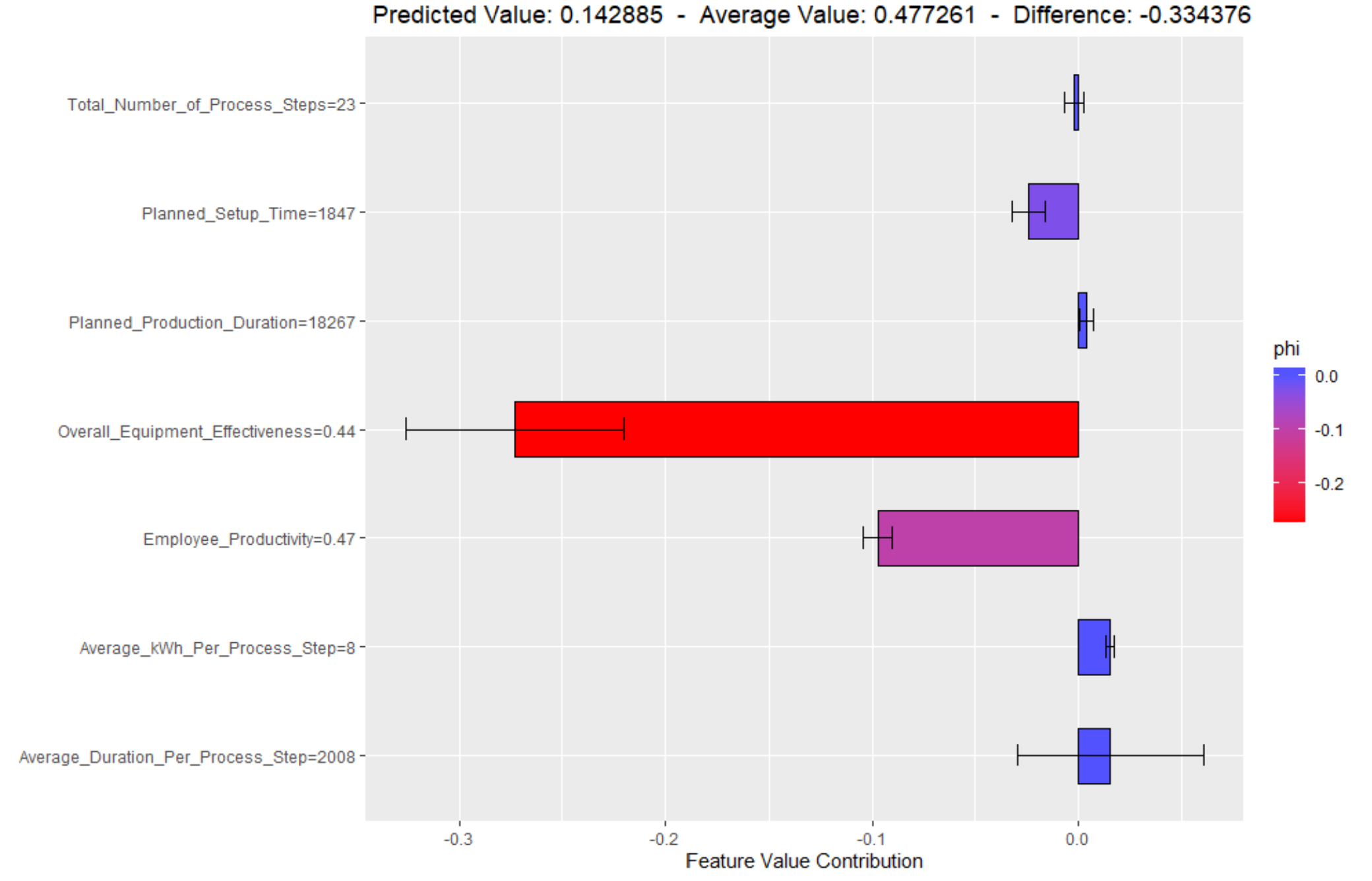}
 \caption{Example of a SHAP-Explanation as it is implemented by Mehdiyev \& Fettke \cite{Mehdiyev2020}, illustrating feature impact on the predictions score using bars, with their length and color representing the contribution of the corresponding feature. The specific feature values as well as the numerical value of their contribution are visible on the axes, the prediction score, the average prediction score and the difference due to feature impact are displayed at the top of the plot.}
 \label{shapley_example_Mehdiyev_figure}
 \end{figure}

\begin{figure}[h]
 \centering
 \includegraphics[width=9cm]{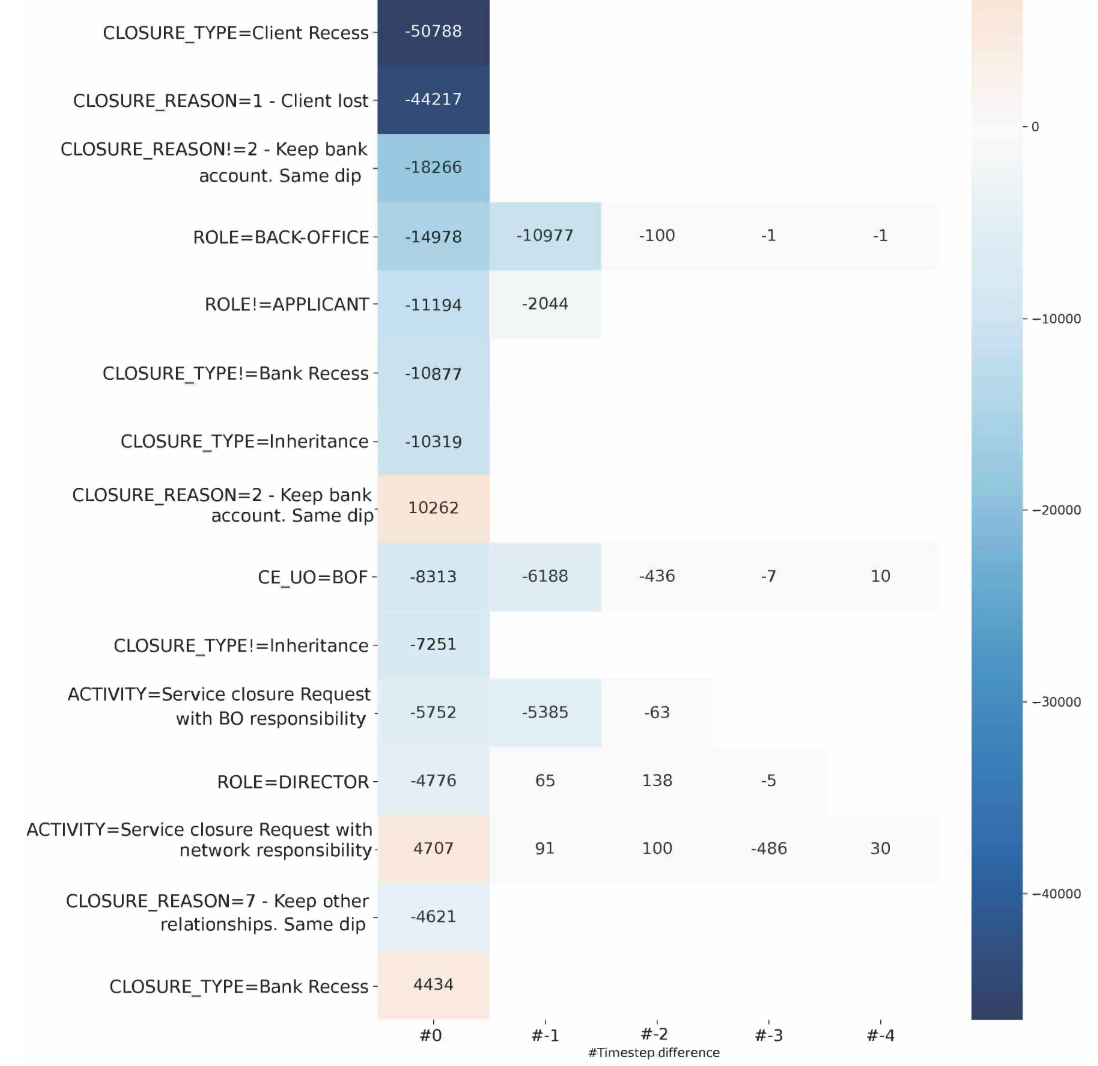}
 \caption{Example of a Shapley-based global explanation as it is implemented by Galanti et al. \cite{Galanti2020}, illustrating frequency of features and corresponding values when they were significantly relevant for the prediction by using a heatmap. Visually, the numeric value for the frequency is being displayed and accentuated via a color gradient: red for positive, blue for negative values. Using this heatmap approach, the x-axis can be used to illustrate the distance between the current and upcoming activity via timesteps.}
 \label{shapley_example_Galanti_figure}
 \end{figure}

\begin{figure}[h]
\centering
\begin{subfigure}{.4\textwidth}
  \centering
  \includegraphics[width=.9\linewidth]{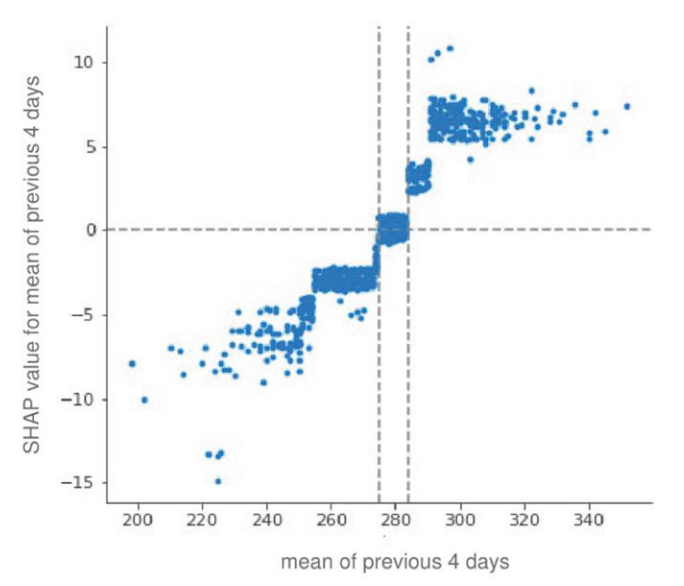}
  \caption{}
  \label{fig:sub1}
\end{subfigure}%
\begin{subfigure}{.6\textwidth}
  \centering
  \includegraphics[width=.9\linewidth]{G15a.PNG}
  \caption{}
  \label{fig:sub2}
\end{subfigure}
\caption{Example for a (a) SHAP Dependence Plot and (b) SHAP Summary Plot as it is implemented by Petsis et al. \cite{Petsis2022}.
(a) illustrates the distribution of SHAP-Values across the scored data set for a specific variable, in very similar fashion to PDPs.
(b) depicts the distribution of SHAP-Values for a subset of feature, arranged by their Feature Importance (y-axis), using colored dots to represent individual instances. The color gradient used for each dot represents the normalized feature value (red implies a high value, blue a low value) and the dot's position represents its SHAP-Value as is visible on the x-axis, while jitter along the y-axis illustrates the distribution of SHAP-Values for the corresponding feature.}
\label{Shaply_global_examples}
\end{figure}

\noindent\textbf{Shapley-based Local and Global Explanations}\\
Shapley values \cite{shapley:book1952} provide a model-agnostic approach, originating from coalition game theory, and illustrate the contribution of individual players towards the final shared profit. For local explanations of ML models, input variables for the model can be considered such players, with the prediction score being the final payout, which is influenced by the feature attributes. Since Shapley values are calculated using all possible coalitions, this method by itself faces the problems of exponential growth during calculation; hence, various implementations exist to circumvent this by using approximations and estimations: SHapley Additive exPlanations (SHAP) (Lundberg et al. \cite{SHAP_Lundberg}) in general as well as model-specific implementations like Kernel SHAP, Linear SHAP, Deep SHAP, etc. provide a method for ML-models to calculate local explanations for specific instances, illustrating the Shapley contribution and, therefore, the impact of certain variables on the prediction score. Local SHAP can also be leveraged to capture global behavior, as demonstrated by Galanti et al. \cite{Galanti2020} in Figure \ref{shapley_example_Galanti_figure} and by Petsis et al. \cite{Petsis2022} in Figure \ref{Shaply_global_examples}, Prominent applications are SHAP Summary Plots (illustrating the distribution of SHAP values for each variable for the whole scored data set) and SHAP Dependence Plots (similar approach as PDP, utilizing Shapley values as a metric for the impact of a variable on the final prediction score). In the analyzed literature, the following articles employed at least one Shapley-based explanation technique: Coma-Puig \& Carmona \cite{coma2022non}, Galanti et al. \cite{Galanti2020}, Galanti et al. \cite{Galanti2022}, Mayer, Mehdiyev \& Fettke \cite{mayer2021manufacturing}, Mehdiyev \& Fettke \cite{Mehdiyev2020a}, Mehdiyev \& Fettke (2020c) \cite{Mehdiyev2020} (Figure \ref{shapley_example_Mehdiyev_figure}), Mehdiyev et al. \cite{mehdiyev2021explainable}, Padella et al. \cite{padella2022explainable}, Petsis et al. \cite{Petsis2022}, Rizzi et al. \cite{Rizzi2020}, Stevens \& de Smedt \cite{Stevens2022}, Stevens et al. \cite{stevens2022assessing}, Stevens et al. \cite{Stevens2022a}, Toh et al. \cite{toh2022improving}, Velmurugan et al. (2021a) \cite{Velmurugan2021a}, Velmurugan et al. (2021b) \cite{Velmurugan2021} and Zilker et al. \cite{zilker2023best}. 

The advantages of Shapley-based approaches lie in the comprehensible distribution of feature contributions towards the final prediction score as well as a solid theoretical foundation that is grounded in game-theory. Furthermore, the reference point for these explanations can be set to specific subsets of the underlying data set, increasing the applicability of this approach to various use cases.\\

\noindent\textbf{Other Local Explanation Methods}\\
In the case of LSTMs, Layerwise Relevance Propagation (LRP) (Lapuschkin et al. \cite{layerwise_relevance_propagation_LRP} and Arras et al. \cite{LRP_for_LSTM}) is a local, model-specific approach that reveals the impact of each feature on the prediction score for a specific instance, as demonstrated by Harl et al. \cite{Harl2020}, Sindghatta et al. (2020b) \cite{Sindhgatta2020a}, Stevens et al. \cite{Stevens2022a}, Weinzierl et al. \cite{Weinzierl2020}, Wickramanayake et al. (2022a) \cite{Wickramanayake2022} and Wickramanayake et al. (2022b) \cite{Wickramanayake2022a}. Although LRP is being presented in this section as a local XAI method, within their articles, Sindghatta et al. (2020b) \cite{Sindhgatta2020a}, and Stevens et al. \cite{Stevens2022a} only presented global explanations on the basis of this method. Similar to LRP, Hanga et al. \cite{Hanga2020} employed a model-specific approach for LSTMs in the context of next-event prediction that allocates probability scores to the possible predicted events. Specifically, for an unfinished trace, the model aims to predict the most likely finishing process trace by encoding the process trace as a graph and displaying the estimated probabilities for each predicted activity. Although this method gives users a confidence score concerning the prediction, the interpretation of these probabilities depends heavily on the use case. Further, this approach misses out on explaining how the estimated probability values came to be. De Koninck et al. \cite{DeKoninck2017} employ SECPI, an approach that trains a Support Vector Machines (SVM), which is inherently not an interpretable model, to identify the minimum number of characteristics a trace can have to remain in the cluster to which it was allocated. This approach primarily focuses on providing explanations for the employed clustering method. The authors define "explainable" instances in their approach as "instances for which such an explanation can be extracted from the underlying SVM"—a highly debatable statement. Huang et al. \cite{Huang2022} present LORELEY, an approach based on LORE (Guidotti et al. \cite{LORE_Guidotti}), which is similar to LIME in that it creates local explanations by training a decision tree within said locality, aiming at capturing local model behavior. LORELEY extends LORE to be applied to predictive process monitoring by modifying the algorithms for calculating trace similarity and distance and for clustering traces. Due to the decision tree as a surrogate model, these types of explanations can also be employed as counterfactual explanations.

While local explanations zoom in on individual predictions, global explanations aim at describing interdependence and relationships between variable expressions and model predictions on a general level, giving insight about the underlying data as well as the model that was trained on said data. Global explanations enable the assessment of the general model behaviour by domain experts and allow for uncovering discrepancies between model behaviour and domain knowledge. The following section presents and illustrates the prevalent global explanation methods among the retrieved articles.\\

\begin{figure}[h]
 \centering
 \includegraphics[width=8cm]{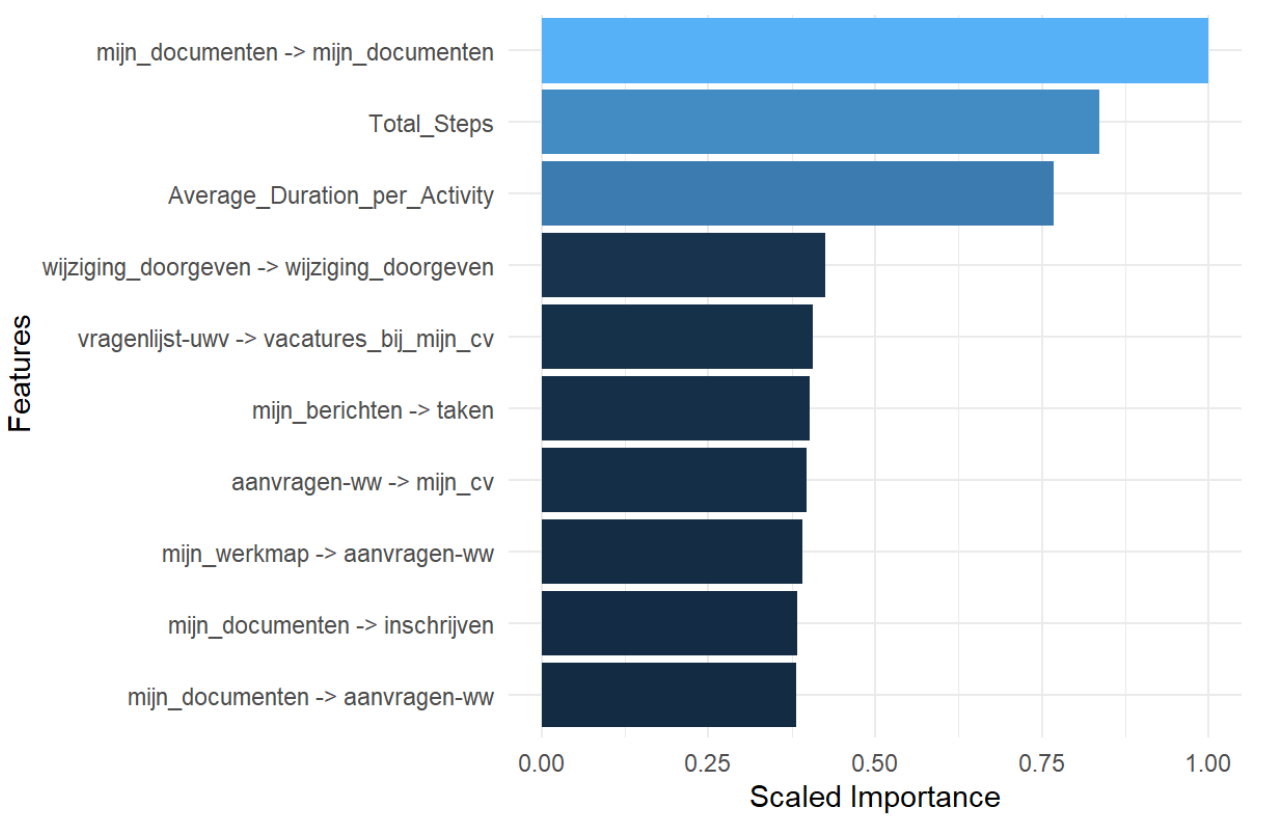}
 \caption{Example of a Feature Importance visualization}
 \label{feature_importance_example_figure}
 \end{figure}

\noindent\textbf{Feature Importance}\\
Feature importance (Gevrey et al. \cite{Feature_Importance_Gevrey}, McDermid \cite{Feat_Imp_and_Example_Based_McDermid} ) is an umbrella term for some of the most prevalent explanation methods observed in the analyzed literature with the objective of identifying the influence of certain features on the calculation of the prediction score. Various feature importance implementations have been observed and although these methods provide viable insight on global characteristics of a given model, some implementations allow for local explanations as well: For permutation feature importance \cite{Perm_Feature_Importance_Fisher}, values of a feature within the data set are being shuffled throughout its instances, then the data set is being re-scored and the mean error is being documented. This process is repeated for any given variable, establishing a ranking of the most influential features, although feature interactions are not captured using this approach. Ouyang et al. \cite{Ouyang2021}, Sindghatta et al. (2020a) \cite{Sindhgatta2020}, Stevens \& de Smedt \cite{Stevens2022},  Stevens et al. \cite{Stevens2022a} implemented the Permutation Feature Importance approach. In the case of LSTMs, feature importance can also be calculated by implementing layerwise relevance propagation (LRP) by averaging relevance scores for each variable over the scored data set, as demonstrated by Harl et al. \cite{Harl2020}, Sindghatta et al. (2020b) \cite{Sindhgatta2020a}, Stevens et al. \cite{Stevens2022a}, Weinzierl et al. \cite{Weinzierl2020}, Wickramanayake et al. (2022a) \cite{Wickramanayake2022} and Wickramanayake et al. (2022b) \cite{Wickramanayake2022a}. Another viable method is leaving out a feature and measuring the model performance after re-training in the style of Feng et al. \cite{LOFO_Feng}, as was done in Bukhsh et al. \cite{AllahBukhsh2019}. Galanti et al. (2022) cite Galanti2022 and Stevens et al. cite Stevens2022a use SHAP feature importance and SHAP Summary Plots, which average local SHAP values of any given variable over the scored data set, as another method to show the impact of variable expressions on the final prediction score. For DFNNs, calculating feature importance based on Gedeon \cite{Feature_Importance_Gedeon} and leveraging connection weights have been employed by Mehdiyev \& Fettke (2020a) \cite{Mehdiyev2020a} and Rehse et al. \cite{Rehse2019} and present another viable approach to uncovering global model behavior. For tree-based models, e.g., XGBoost as in Stevens et al. \cite{Stevens2022a}, the mean impact on Gini-index-based purity for each feature can be leveraged in order to calculate Feature Importance as well. Figure \ref{feature_importance_example_figure} is an exemplary visualization of feature importance from Mehdiyev \& Fettke (2020a) \cite{Mehdiyev2020a}, depicting the scaled importance of the ten most significant features via a bar plot, with the length and coloration of each bar representing the impact the feature has on the calculation of the prediction score.\\

\begin{figure}[h]
 \centering
 \includegraphics[width=8cm]{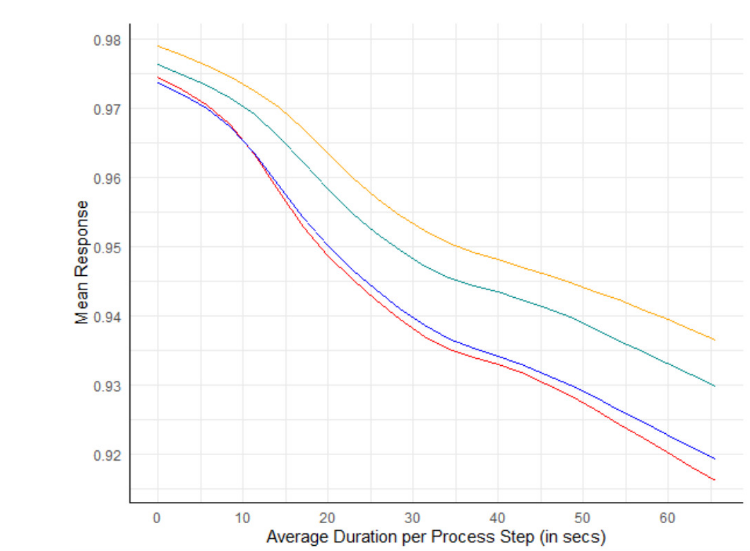}
 \caption{Example of a Partial Dependence Plot}
 \label{pdp_example_figure}
 \end{figure}

\noindent\textbf{Partial Dependence Plot (PDP)}\\
A PDP \cite{PDP_Friedman} illustrates the impact of feature expressions of a given variable on the prediction score, although it does not capture the influence of and on other features. The basic principle behind this method is the iterative re-scoring of the data set after permuting the value of a chosen variable. The PDP value of a variable at a certain variable expression captures the average prediction score of the corresponding data set if the chosen variable was set to the said expression for each instance within the data set. This way, the impact on the prediction score of marginal changes in the variable expression can be captured, allowing for the validation of the model decision-making by domain experts. Although this method is easy to interpret, feature interdependencies cannot be revealed using this method alone; in such a case, the corresponding PDP might be misleading to the user. Furthermore, for categorical variables, the amount of permutations increases quadratically, and the same is true for numeric variables, given that not only samples but any observed feature value is being used for the permutations. Figure \ref{pdp_example_figure} is an example of a PDP from Mehdiyev \& Fettke (2020a) \cite{Mehdiyev2020a}, illustrating the mean prediction score based on the value of the variable "Average Duration per Process Step", with each colored line representing an age group. It is visible that the average response decreases with increased duration per process step, with age being a significant contributing factor to the prediction score as well.

\begin{table}[p]
\begin{center} 
\begin{minipage}{330pt}
\setlength{\tabcolsep}{2pt}
\caption{Categorization of employed ML and explanation methods in the found literature, segmented into model interpretabilty, explanation scope, explanation relation and explanation format.}
\label{table_model_expl_A}
\begin{tabular}{c|cccc|cccc|cccccc|cccc|cc|cccc}
\toprule
& \multicolumn{4}{c|}{\footnotesize Interpret-}      & \multicolumn{20}{c}{\footnotesize Explainable AI}\\ 
& \multicolumn{4}{c|}{\footnotesize able AI}         & \multicolumn{20}{c}{\footnotesize }\\ 

& \multicolumn{4}{c|}{\footnotesize }  & \multicolumn{4}{c|}{\footnotesize }  & \multicolumn{10}{c|}{\footnotesize Scope}  & \multicolumn{2}{c|}{\footnotesize Rel-}   & \multicolumn{4}{c}{\footnotesize Format}\\
& \multicolumn{4}{c|}{\footnotesize }  & \multicolumn{4}{c|}{\footnotesize }  & \multicolumn{10}{c|}{\footnotesize}  & \multicolumn{2}{c|}{\footnotesize ation}  & \multicolumn{4}{c}{\footnotesize }\\

& \multicolumn{4}{c|}{\footnotesize White-Box} & \multicolumn{4}{c|}{\footnotesize Black-Box}            & \multicolumn{6}{c|}{\footnotesize Local} & \multicolumn{4}{c|}{\footnotesize Global} & \multicolumn{2}{c|}{\footnotesize } & \multicolumn{4}{c}{\footnotesize}\\

Publication &
\rot{\footnotesize Bayesian Network} & \rot{\footnotesize Decision Tree} & \rot{\footnotesize Lin./Log. Regression} & \rot{\footnotesize Other} & 
\rot{\footnotesize Deep Learning} & \rot{\footnotesize Gradient Boosting} & \rot{\footnotesize Random Forest} & \rot{\footnotesize Other} & 
\rot{\footnotesize Counterfactual} & \rot{\footnotesize Feature Importance} & \rot{\footnotesize ICE} & \rot{\footnotesize LIME} & \rot{\footnotesize Shapley-based} & \rot{\footnotesize Other} &  
\rot{\footnotesize Feature Importance} & \rot{\footnotesize Shapley-based} & \rot{\footnotesize PDP} & \rot{\footnotesize Other} &  
\rot{\footnotesize Model-agnostic} & \rot{\footnotesize Model-specific} & 
\rot{\footnotesize Numeric} & \rot{\footnotesize Rule-based} & \rot{\footnotesize Textual} & \rot{\footnotesize Visual} \\
\botrule
\end{tabular}
\end{minipage}
\end{center}
\end{table}

\begin{table}[p]
\begin{center} 
\begin{minipage}{330pt}
\setlength{\tabcolsep}{2pt}
\caption{Categorization of employed ML and explanation methods in the found literature, segmented into model interpretabilty, explanation scope, explanation relation and explanation format.}
\label{table_model_expl_B}
\begin{tabular}{c|cccc|cccc|cccccc|cccc|cc|cccc}
\toprule
& \multicolumn{4}{c|}{\footnotesize Interpret-}      & \multicolumn{20}{c}{\footnotesize Explainable AI}\\ 
& \multicolumn{4}{c|}{\footnotesize able AI}         & \multicolumn{20}{c}{\footnotesize }\\ 

& \multicolumn{4}{c|}{\footnotesize }  & \multicolumn{4}{c|}{\footnotesize }  & \multicolumn{10}{c|}{\footnotesize Scope}  & \multicolumn{2}{c|}{\footnotesize Rel-}   & \multicolumn{4}{c}{\footnotesize Format}\\
& \multicolumn{4}{c|}{\footnotesize }  & \multicolumn{4}{c|}{\footnotesize }  & \multicolumn{10}{c|}{\footnotesize}  & \multicolumn{2}{c|}{\footnotesize ation}  & \multicolumn{4}{c}{\footnotesize }\\

& \multicolumn{4}{c|}{\footnotesize White-Box} & \multicolumn{4}{c|}{\footnotesize Black-Box}            & \multicolumn{6}{c|}{\footnotesize Local} & \multicolumn{4}{c|}{\footnotesize Global} & \multicolumn{2}{c|}{\footnotesize } & \multicolumn{4}{c}{\footnotesize}\\

Publication &
\rot{\footnotesize Bayesian Network} & \rot{\footnotesize Decision Tree} & \rot{\footnotesize Lin./Log. Regression} & \rot{\footnotesize Other} & 
\rot{\footnotesize Deep Learning} & \rot{\footnotesize Gradient Boosting} & \rot{\footnotesize Random Forest} & \rot{\footnotesize Other} & 
\rot{\footnotesize Counterfactual} & \rot{\footnotesize Feature Importance} & \rot{\footnotesize ICE} & \rot{\footnotesize LIME} & \rot{\footnotesize Shapley-based} & \rot{\footnotesize Other} &  
\rot{\footnotesize Feature Importance} & \rot{\footnotesize Shapley-based} & \rot{\footnotesize PDP} & \rot{\footnotesize Other} &  
\rot{\footnotesize Model-agnostic} & \rot{\footnotesize Model-specific} & 
\rot{\footnotesize Numeric} & \rot{\footnotesize Rule-based} & \rot{\footnotesize Textual} & \rot{\footnotesize Visual} \\
\botrule
\end{tabular}
\end{minipage}
\end{center}
\end{table}

\newpage
\subsection{Evaluation of Explainability and Interpretability}\label{explanation_evaluation}
The evaluation of explainability and interoperability in ML is a complex endeavor, requiring a nuanced understanding of different methodologies, each with its unique strengths and considerations. This section delves into the comparative analysis of quantitative versus qualitative evaluation methods and explores the multifaceted approach of functional, application, and human-grounded evaluations \cite{doshi2017towards}.

\subsubsection{Indvidiual Studies}

In the analyzed literature, the evaluation of proposed XAI-methods varied with characteristics of the underlying method, its users and goals, the model in need of explanations as well as the application context. This section presents the evaluation methods of the analyzed articles (see Tables \ref{table_expl_eva_A} and \ref{table_expl_eva_B}).

De Koninck et al. \cite{DeKoninck2017} evaluate their implementation of SECPI by comparing the runtime in seconds, the length of explanations, i.e. the number of created rules that explain why an instance belongs to a specific cluster, as well as the relative amount of "explainable" instances, i.e. the relative amount of instance for which the employed SVM was able to find minimal sets of rules that allow the instance to stay in its allocated cluster.

Folino et al. \cite{Folino2017} evaluate their approach for extracting explanations for trace clustering by providing clustering rules on "explanation complexity", i.e. the number of rules needed to justify a traces allocation to a specific cluster, as well as interestingness and compared the results to an explainable M5Rules (Holmes et al. \cite{m5rules_holmes}) implementation.

Galanti et al. \cite{Galanti2022} employ a two-parted approach to evaluating their utilized explanation approach: First, explanations are evaluated on their soundness based on statistical analysis and domain knowledge. Second, a user-evaluation with 20 participants has been conducted, with the participants solving 18 tasks and reporting their personal estimation of the difficulty of said tasks. Afterwards, usability and user experience have been captured using questionnaires.

Hsieh et al. \cite{Hsieh2021} evaluate the quality of their counterfactual explanations with regards to diversity, plausibility, proximity, sparsity and whether the explanations can incorporate categorical features. In this context, diversity refers to the amount of different counterfactual explanations created,
plausibility refers to the soundness of the counterfactual explanations based on domain knowledge, proximity refers to the proximity of the counterfactual explanations and the instance given as input based on the distance measurement, sparsity refers to the mean amount of modified features that constitute a counterfactual explanation for the instance given as an input. The evaluation incorporates a statistical approach as well as the evaluation of explanations for specific traces.

Mehdiyev \& Fettke (2020b) \cite{Mehdiyev2021} used the coefficient of determination (R²-value) for the surrogate model for each locality in order to reveal the quality of the surrogate capturing the behaviour of the underlying model. Due to the surrogate models being inherently interpretable Decision Trees, the provided explanations were not evaluated individually.

Stevens \& de Smedt \cite{Stevens2022} evaluate their employed XAI-methods with regards to functional complexity, level of disagreement and parsimony, : For the authors, in this context,
functional complexity refers to a  metric, similar to the measurement of permutated feature importance, that captures how easily a prediction can be manipulated when altering certain feature values,
level of disagreement (Lakkaraju et al. \cite{level_of_disagreement_Lakkaraju}) refers to discrepancies with regards to the prediction score between the underlying model and corresponding surrogate models, and parsimony refers to the trade-off between the simplicity of provided explanations and the performance, i.e. accuracy, of the underlying model.

Velmurugan et al. (2021a) \cite{Velmurugan2021a} differentiate in their evaluation of XAI-methods internal \& external fidelity, refering to the definition of fidelity from Messalas et al. \cite{fidelity_Messalas}: External fidelity measures the similarity between the predictions of the underlying model and corresponding surrogate model, whereas internal fidelity focuses on the decision-making process of the models, specifically on the amount of similarities between these models. The authors focused on the internal fidelity of LIME and SHAP and for its measurement, instances were perturbed ten times and the mean absolute percentage error between the task model and surrogate model was documented.

Velmurugan et al. (2021b) \cite{Velmurugan2021} evaluated the stability, refering to Visani et al. \cite{stability_Visani}, aiming at measuring the constistency of explanations for same or similar instances. In particular, the stability of the identified most important features (a subgroup of feature residing in the top quartile with regards to the weight distribution) as well as the stability of corresponding weights was examined. The authors used this approach to evaluate the employed LIME and SHAP methods. 

\begin{table}[p]
\begin{center} 
\begin{minipage}{210pt}
\setlength{\tabcolsep}{2pt}
\caption{Categorization of employed explanation evaluation methods and metrics in the found literature.}
\label{table_expl_eva_A}
\begin{tabular}{c|cc|cc|ccc|cccc}
\toprule
& \multicolumn{11}{c}{\footnotesize Evaluation}\\ 

& \multicolumn{2}{c|}{\footnotesize Performed}  & \multicolumn{2}{c|}{\footnotesize Type}    & \multicolumn{3}{c|}{\footnotesize Method}   & \multicolumn{4}{c}{\footnotesize Metrics}\\ 

Publication &
No & Yes &  
\rot{\footnotesize Qualitative} & \rot{\footnotesize Quantitative} &  
\rot{\footnotesize Application-grounded} & \rot{\footnotesize Functional-grounded} & \rot{\footnotesize Human-grounded} &  
\rot{\footnotesize Fidelity} & \rot{\footnotesize Functional Complexity} & \rot{\footnotesize Parsimony} & \rot{\footnotesize Stability}\\
\hline

\footnotesize Agarwal et al. \cite{agarwal2022process}  & 
$\blacksquare$ & 
& 

& 
& 

& 
& 
& 

& 
& 
& 

\\
\hline

\footnotesize Bayomie et al. \cite{bayomie2022improving} & 
$\blacksquare$ & 
& 

& 
& 

& 
& 
& 

& 
& 
& 

\\
\hline

\footnotesize  Böhmer \&  & 
$\blacksquare$ & 
& 

& 
& 

& 
& 
& 

& 
& 
& 

\\ 

\footnotesize Rinderle-Ma \cite{bohmer2018probability}  & 
$\blacksquare$ & 
& 

& 
& 

& 
& 
& 

& 
& 
& 

\\
\hline

\footnotesize Böhmer \&  & 
& 
$\blacksquare$ & 

$\blacksquare$ & 
& 

& 
& 
$\blacksquare$ & 

& 
& 
& 

\\

\footnotesize Rinderle-Ma \cite{bohmer2020mining} & 
& 
& 

& 
& 

& 
& 
& 

& 
& 
& 

\\
\hline

\footnotesize  Böhmer \& & 
$\blacksquare$ & 
& 

& 
& 

& 
& 
& 

& 
& 
& 

\\

\footnotesize  Rinderle-Ma \cite{Bohmer2020}  & 
& 
& 

& 
& 

& 
& 
& 

& 
& 
& 

\\
\hline

\footnotesize Brunk et al. \cite{Brunk2021}  & 
$\blacksquare$ & 
& 

& 
& 

& 
& 
& 

& 
& 
& 

\\
\hline

\footnotesize  Bukhsh et al. \cite{AllahBukhsh2019}     & 
$\blacksquare$ & 
& 

& 
& 

& 
& 
& 

& 
& 
& 

\\
\hline

\footnotesize Cao et al. \cite{cao2023business}  & 
$\blacksquare$ & 
& 

& 
& 

& 
& 
& 

& 
& 
& 

\\
\hline

\footnotesize Cao et al. \cite{cao2023explainable} & 
$\blacksquare$ & 
& 

& 
& 

& 
& 
& 

& 
& 
& 

\\
\hline

\footnotesize Coma-Puig \&  & 
$\blacksquare$ & 
& 

& 
& 

& 
& 
& 

& 
& 
& 

\\

\footnotesize Carmona \cite{coma2022non}   & 
& 
& 

& 
& 

& 
& 
& 

& 
& 
& 

\\
\hline

\footnotesize  Conforti et al. \cite{Conforti2016} & 
$\blacksquare$ & 
& 

& 
& 

& 
& 
& 

& 
& 
& 

\\
\hline

\footnotesize De Koninck et al. \cite{DeKoninck2017}  & 
& 
$\blacksquare$ & 

& 
$\blacksquare$ & 

$\blacksquare$ & 
& 
& 

& 
& 
$\blacksquare$ & 

\\
\hline

\footnotesize De Leoni et al. \cite{DeLeoni2015}  & 
$\blacksquare$ & 
& 

& 
& 

& 
& 
& 

& 
& 
& 

\\
\hline

\footnotesize De Oliveira et al. \cite{de2020optimization}  & 
$\blacksquare$ & 
& 

& 
& 

& 
& 
& 

& 
& 
& 

\\
\hline

\footnotesize De Oliveira et al. \cite{de2020automatic}  & 
$\blacksquare$ & 
& 

& 
& 

& 
& 
& 

& 
& 
& 

\\
\hline

\footnotesize Di Francescomarino  & 
$\blacksquare$ & 
& 

& 
& 

& 
& 
& 

& 
& 
& 

\\

\footnotesize et al. \cite{DiFrancescomarino2016}  & 
& 
& 

& 
& 

& 
& 
& 

& 
& 
& 

\\
\hline

\footnotesize Di Francescomarino & 
$\blacksquare$ & 
& 

& 
& 

& 
& 
& 

& 
& 
& 

\\

\footnotesize et al. \cite{Francescomarino2019}  & 
& 
& 

& 
& 

& 
& 
& 

& 
& 
& 

\\
\hline

\footnotesize Folino et al. \cite{Folino2017}  & 
& 
$\blacksquare$ & 

$\blacksquare$ & 
& 

$\blacksquare$ & 
& 
& 

& 
& 
$\blacksquare$ & 

\\
\hline

\footnotesize Fu et al. \cite{fu2021modeling} & 
$\blacksquare$ & 
& 

& 
& 

& 
& 
& 

& 
& 
& 

\\
\hline

\footnotesize Galanti et al. \cite{Galanti2020}  & 
$\blacksquare$ & 
& 

& 
& 

& 
& 
& 

& 
& 
& 

\\
\hline

\footnotesize Galanti et al. \cite{Galanti2022}  & 
& 
$\blacksquare$ & 

$\blacksquare$ & 
$\blacksquare$ & 

& 
& 
$\blacksquare$ & 

& 
& 
$\blacksquare$ & 

\\
\hline

\footnotesize Garcia-Banuelos  & 
& 
$\blacksquare$ & 

$\blacksquare$ & 
$\blacksquare$ & 

$\blacksquare$ & 
& 
$\blacksquare$ & 

& 
& 
$\blacksquare$ & 

\\ 

\footnotesize  et al. \cite{garcia2017complete}  & 
& 
& 

& 
& 

& 
& 
& 

& 
& 
& 

\\
\hline

\footnotesize Gerlach et al. \cite{gerlach2022inferring} & 
$\blacksquare$ & 
& 

& 
& 

& 
& 
& 

& 
& 
& 

\\
\hline

\footnotesize Hanga et al. \cite{Hanga2020} & 
$\blacksquare$ & 
& 

& 
& 

& 
& 
& 

& 
& 
& 

\\
\hline

\footnotesize Harl et al. \cite{Harl2020}   & 
$\blacksquare$ & 
& 

& 
& 

& 
& 
& 

& 
& 
& 

\\
\hline

\footnotesize Horita et al. \cite{horita2016goal}  & 
$\blacksquare$ & 
& 

& 
& 

& 
& 
& 

& 
& 
& 

\\
\hline

\footnotesize Hsieh et al. \cite{Hsieh2021}  & 
$\blacksquare$ & 
& 

& 
& 

& 
& 
& 

& 
& 
& 

\\
\hline

\footnotesize  Huang et al. \cite{Huang2022}  & 
$\blacksquare$ & 
& 

& 
& 

& 
& 
& 

& 
& 
& 

\\
\hline

\footnotesize Irarrazaval et al. \cite{irarrazaval2021telecom}  & 
$\blacksquare$ & 
& 

& 
& 

& 
& 
& 

& 
& 
& 

\\
\hline

\footnotesize Khemiri \& Pinaton \cite{khemiri2018improving}  & 
$\blacksquare$ & 
& 

& 
& 

& 
& 
& 

& 
& 
& 

\\
\hline

\footnotesize Lakshmanan et al. \cite{Lakshmanan2011}  & 
$\blacksquare$ & 
& 

& 
& 

& 
& 
& 

& 
& 
& 

\\
\hline

\footnotesize  Maggi et al. \cite{Maggi2014}  & 
$\blacksquare$ & 
& 

& 
& 

& 
& 
& 

& 
& 
& 

\\
\hline

\footnotesize Mayer, Mehdiyev \&   & 
$\blacksquare$ & 
& 

& 
& 

& 
& 
& 

& 
& 
& 

\\ 

\footnotesize  Fettke \cite{mayer2021manufacturing}  & 
& 
& 

& 
& 

& 
& 
& 

& 
& 
& 

\\

\botrule
\end{tabular}
\end{minipage}
\end{center}
\end{table}

\begin{table}[p]
\begin{center} 
\begin{minipage}{230pt}
\setlength{\tabcolsep}{2pt}
\caption{Categorization of employed explanation evaluation methods and metrics in the found literature.}
\label{table_expl_eva_B}
\begin{tabular}{c|cc|cc|ccc|cccc}
\toprule
& \multicolumn{11}{c}{\footnotesize Evaluation}\\ 

& \multicolumn{2}{c|}{\footnotesize Performed}  & \multicolumn{2}{c|}{\footnotesize Type}    & \multicolumn{3}{c|}{\footnotesize Method}   & \multicolumn{4}{c}{\footnotesize Metrics}\\ 

Publication &
No & Yes &  
\rot{\footnotesize Qualitative} & \rot{\footnotesize Quantitative} &  
\rot{\footnotesize Application-grounded} & \rot{\footnotesize Functional-grounded} & \rot{\footnotesize Human-grounded} &  
\rot{\footnotesize Fidelity} & \rot{\footnotesize Functional Complexity} & \rot{\footnotesize Parsimony} & \rot{\footnotesize Stability}\\
\hline

\footnotesize Mehdiyev \& Fettke \cite{Mehdiyev2021} & 
& 
$\blacksquare$ & 

& 
$\blacksquare$ & 

\phantom{$\blacksquare$} & 
$\blacksquare$ & 
& 

$\blacksquare$ & 
& 
& 

\\
\hline

\footnotesize   Mehdiyev \& Fettke \cite{Mehdiyev2020a}   & 
$\blacksquare$ & 
& 

& 
& 

& 
& 
& 

& 
& 
& 

\\
\hline

\footnotesize  Mehdiyev \& Fettke \cite{Mehdiyev2020}   & 
$\blacksquare$ & 
& 

& 
& 

& 
& 
& 

& 
& 
& 

\\
\hline

\footnotesize   Mehdiyev et al. \cite{mehdiyev2021explainable}   & 
$\blacksquare$ & 
& 

& 
& 

& 
& 
& 

& 
& 
& 

\\
\hline

\footnotesize  Ouyang et al. \cite{Ouyang2021}    & 
$\blacksquare$ & 
& 

& 
& 

& 
& 
& 

& 
& 
& 

\\
\hline

\footnotesize  Padella et al. \cite{padella2022explainable}  & 
$\blacksquare$ & 
& 

& 
& 

& 
& 
& 

& 
& 
& 

\\
\hline

\footnotesize  Pasquadibisceglie  & 
$\blacksquare$ & 
& 

& 
& 

& 
& 
& 

& 
& 
& 

\\ 

\footnotesize  et al. \cite{Pasquadibisceglie2021}   & 
$\blacksquare$ & 
& 

& 
& 

& 
& 
& 

& 
& 
& 

\\
\hline

\footnotesize   Pauwels \& Calders \cite{pauwels2019detecting}  & 
$\blacksquare$ & 
& 

& 
& 

& 
& 
& 

& 
& 
& 

\\
\hline

\footnotesize  Pauwels \& Calders \cite{pauwels2019anomaly}   & 
$\blacksquare$ & 
& 

& 
& 

& 
& 
& 

& 
& 
& 

\\
\hline

\footnotesize  Petsis et al. \cite{Petsis2022}    & 
$\blacksquare$ & 
& 

& 
& 

& 
& 
& 

& 
& 
& 

\\
\hline

\footnotesize  Polato et al. \cite{polato2018time}    & 
$\blacksquare$ & 
& 

& 
& 

& 
& 
& 

& 
& 
& 

\\
\hline

\footnotesize  Prasisdis et al. \cite{prasidis2021handling}   & 
$\blacksquare$ & 
& 

& 
& 

& 
& 
& 

& 
& 
& 

\\
\hline

\footnotesize  Rehse et al. \cite{Rehse2019}  & 
$\blacksquare$ & 
& 

& 
& 

& 
& 
& 

& 
& 
& 

\\
\hline

\footnotesize Rizzi et al. \cite{Rizzi2020}    & 
$\blacksquare$ & 
& 

& 
& 

& 
& 
& 

& 
& 
& 

\\
\hline

\footnotesize  Savickas \& Vasilecas \cite{savickas2014business}   & 
$\blacksquare$ & 
& 

& 
& 

& 
& 
& 

& 
& 
& 

\\
\hline

\footnotesize   Savickas \& Vasilecas \cite{savickas2018belief}   & 
$\blacksquare$ & 
& 

& 
& 

& 
& 
& 

& 
& 
& 

\\
\hline

\footnotesize  Sindghatta et al. \cite{Sindhgatta2020}   & 
$\blacksquare$ & 
& 

& 
& 

& 
& 
& 

& 
& 
& 

\\
\hline
  
\footnotesize Sindghatta et al. \cite{Sindhgatta2020a}  & 
$\blacksquare$ & 
& 

& 
& 

& 
& 
& 

& 
& 
& 

\\
\hline

\footnotesize Stevens \&  & 
& 
$\blacksquare$ & 

& 
$\blacksquare$ & 

& 
$\blacksquare$ & 
& 

$\blacksquare$ & 
$\blacksquare$ & 
$\blacksquare$ & 

\\

\footnotesize de Smedt \cite{Stevens2022}   & 
& 
& 

& 
& 

& 
& 
& 

& 
& 
& 

\\
\hline
\footnotesize Stevens et al. \cite{stevens2022assessing}  & 
$\blacksquare$ & 
& 

& 
& 

& 
& 
& 

& 
& 
& 

\\
\hline
\footnotesize  Stevens et al. \cite{Stevens2022a}  & 
$\blacksquare$ & 
& 

& 
& 

& 
& 
& 

& 
& 
& 

\\
\hline
\footnotesize  Tama et al.\cite{tama2020empirical}  & 
$\blacksquare$ & 
& 

& 
& 

& 
& 
& 

& 
& 
& 

\\
\hline
\footnotesize   Teinemaa et al. \cite{Teinemaa2016} & 
$\blacksquare$ & 
& 

& 
& 

& 
& 
& 

& 
& 
& 

\\
\hline
\footnotesize  Toh et al. \cite{toh2022improving} & 
$\blacksquare$ & 
& 

& 
& 

& 
& 
& 

& 
& 
& 

\\
\hline
\footnotesize  Unuvar et al. \cite{Unuvar2016}  & 
$\blacksquare$ & 
& 

& 
& 

& 
& 
& 

& 
& 
& 

\\
\hline
\footnotesize Velmurugan et al. \cite{Velmurugan2021a}   & 
& 
$\blacksquare$ & 

& 
$\blacksquare$ & 

& 
$\blacksquare$ & 
& 

$\blacksquare$ & 
& 
& 

\\
\hline
\footnotesize Velmurugan et al. \cite{Velmurugan2021}   & 
& 
$\blacksquare$ & 

& 
$\blacksquare$ & 

& 
$\blacksquare$ & 
& 

& 
& 
& 
$\blacksquare$ 

\\
\hline
\footnotesize  Verenich et al. \cite{Verenich2016}   & 
$\blacksquare$ & 
& 

& 
& 

& 
& 
& 

& 
& 
& 

\\
\hline
\footnotesize Verenich et al. \cite{Verenich2017}   & 
$\blacksquare$ & 
& 

& 
& 

& 
& 
& 

& 
& 
& 

\\
\hline
\footnotesize   Verenich et al. \cite{Verenich2019b}  & 
$\blacksquare$ & 
& 

& 
& 

& 
& 
& 

& 
& 
& 

\\
\hline
\footnotesize   Weinzierl et al. \cite{Weinzierl2020}  & 
$\blacksquare$ & 
& 

& 
& 

& 
& 
& 

& 
& 
& 

\\
\hline
    
\footnotesize Wickramanayake& 
$\blacksquare$ & 
& 

& 
& 

& 
& 
& 

& 
& 
& 

\\ 

\footnotesize et al. \cite{Wickramanayake2022}  & 
$\blacksquare$ & 
& 

& 
& 

& 
& 
& 

& 
& 
& 

\\
\hline

\footnotesize Wickramanayake & 
$\blacksquare$ & 
& 

& 
& 

& 
& 
& 

& 
& 
& 

\\
\hline

\footnotesize et al. \cite{Wickramanayake2022a} & 
$\blacksquare$ & 
& 

& 
& 

& 
& 
& 

& 
& 
& 

\\
\hline

\footnotesize Zilker et al. \cite{zilker2023best}  & 
& 
$\blacksquare$ & 

$\blacksquare$& 
& 

& 
& 
$\blacksquare$ & 

& 
& 
& 

\\ 





 
\botrule
\end{tabular}
\end{minipage}
\end{center}
\end{table}

\subsubsection{Evaluation Type: Quantitative vs. Qualitative Evaluation}

The evaluation of explainability methodologies is a multifaceted task, encompassing the adoption of both qualitative and quantitative methodologies. The significance of quantitative metrics in the evaluation of XAI is emphasized by both Li (2021)  \cite{li2021experimental} and Rosenfeld (2021)\cite{rosenfeld2021better}. Li's research reveals that no single method exhibits superiority across all metrics, underscoring the need for a comprehensive evaluation framework. On the other hand, Rosenfeld proposes four distinct metrics that can be employed to quantify the explanatory nature of XAI systems. Nauta et al. (2022) underscore the imperative of conducting a thorough and all-encompassing evaluation, wherein the author delineates twelve distinct properties that warrant careful assessment \cite{nauta2023anecdotal}. Nevertheless, it is worth noting that anecdotal evidence and user studies are commonly employed in the evaluation of XAI. This observation implies that a comprehensive approach that integrates both qualitative and quantitative methodologies is required \cite{mohseni2021multidisciplinary}.

Of the 67 papers reviewed for XAI in predictive process monitoring, a majority did not engage in any formal evaluation, while only a few employed quantitative or qualitative methods, and even fewer integrated both. This indicates a gap in the current research practices, where the nuances and user-centric aspects crucial for the adoption and trustworthiness of XAI systems might be overlooked. The hypothesis here is that integrating both quantitative and qualitative methods can provide a more holistic understanding of an AI system's explainability, balancing the objectivity of numerical data with the depth of descriptive analysis.

\subsubsection{Evaluation Method: Application, Human and Functional Grounded Methods}

Transitioning from the dichotomy of quantitative and qualitative evaluations, the framework proposed by Doshi-Velez (2017) offers a more granular understanding of XAI evaluation through functional, application, and human-grounded methodologies \cite{doshi2017towards}. Functional grounded evaluation delves into the theoretical and technical soundness of explanations. It's a critical approach for ensuring that the XAI methods align with established cognitive and computational frameworks, as highlighted by \cite{mehdiyev2021explainable}. This approach is vital for the foundational integrity of XAI systems, ensuring that they are not only effective but also theoretically sound.

Application-grounded evaluation shifts the focus to the practical impact of XAI, examining how explainers influence specific decision-making tasks. This methodology is crucial for assessing the real-world utility of XAI, ensuring that the explanations provided are not only understandable but also actionable and beneficial in practical scenarios. Meanwhile, human-grounded evaluation, as discussed by Mohseni (2018) \cite{mohseni2021multidisciplinary}, centers on the user's perspective, measuring how effectively an XAI system's explanations foster trust and understanding among its human users. This approach is paramount for the user-centric development of XAI systems, ensuring that they meet the actual needs and expectations of the people they are designed to assist.

In our study, a balanced exploration across these dimensions was observed, yet the overall engagement in comprehensive evaluation was limited. This indicates a recognition of the importance of diverse evaluative lenses but also hints at the challenges and complexities inherent in implementing such multifaceted methodologies. While the field acknowledges the need for a broad spectrum of evaluation strategies, the practical implementation is still catching up, requiring more robust frameworks and tools to facilitate these comprehensive assessments.

In conclusion, the evaluation of XAI systems is an intricate task, necessitating a balanced and thorough approach that encompasses both quantitative and qualitative methods, as well as functional, application, and human-grounded evaluations. The current research landscape shows a tendency towards quantitative methods and reveals a significant gap in formal evaluation practices. To advance the field of XAI and ensure the development of effective, reliable, and user-centered systems, a more rigorous and holistic approach to evaluation is imperative. As the field continues to evolve, embracing this multifaceted evaluation paradigm will be crucial for the maturation and widespread adoption of explainable and trustworthy AI systems.

\hfill

\section{Discussion}\label{sec6}
\subsection{Challenges and Open Issues}

The critical exploration of explainable and interpretable AI surfaces a multitude of challenges and open issues, pivotal among which is the frequent omission of proper evaluation. A significant proportion of studies in the field prioritize the accuracy of ML algorithms, often relegating the evaluation of explainability and interpretability to a secondary concern. This singular focus not only undermines the core tenet of XAI—making complex algorithms understandable to humans—but also risks the utility of these systems in practical scenarios where understanding the 'why' behind decisions is as important as the decisions themselves.

For those studies that do venture into the evaluation of their XAI approaches, many anchor themselves firmly in either qualitative or quantitative domains. The resultant analyses are thereby one-dimensional, offering a sliver of insight into either the measurable effectiveness or the subjective user experience of the explanations generated. What this dichotomy fails to capture is the nuanced interplay between these two facets in real-world applications. A more comprehensive, multifaceted approach is called for—one that synthesizes both quantitative precision and qualitative depth to yield a richer, more rounded assessment of XAI methods.

The predilection for using benchmark datasets, such as the BPI datasets, exacerbates this issue. These datasets allow for rigorous quantitative analysis, yet they simultaneously constrain the possibility of qualitative assessment due to the lack of access to domain experts. These experts are crucial for interpreting the results within a meaningful context, ensuring that the explanations provided by XAI systems align with domain-specific knowledge and practical realities. Further complicating the landscape is the issue of transferability. The tendency of studies to narrow their focus to specific domains, such as healthcare or finance, begs the question of how well these solutions can be applied across different fields. This siloed approach to research overlooks the importance of generalization properties, leaving unaddressed the potential for XAI solutions to adapt to and function within a variety of domains.

Moreover, the scarcity of real-world studies presents a considerable gap in the literature. The evaluations that do exist often occur in controlled "laboratory" environments, devoid of the economic and organizational contexts that heavily influence the feasibility, scalability, and economic viability of XAI solutions for predictive process monitoring. Without the consideration of these broader factors, the evaluations remain theoretical exercises rather than practical analyses.

In this respect, the discussion points to the necessity for XAI research to transcend its current confines. To advance, it must embrace evaluations that not only traverse the spectrum from quantitative to qualitative but also consider the systemic implications of deploying XAI in diverse, real-world settings. By integrating economic and organizational considerations, future research can aspire to develop XAI solutions that are not only technically robust and understandable but also practically implementable and economically sustainable. Such holistic evaluations will provide a crucial bridge between the theoretical promise of XAI and its real-world applicability, ultimately driving the field towards mature, responsible, and widespread use of interpretable and explainable systems.

\subsection{Practical Implications}

The practical implications of explainability and interpretability in the realm of predictive process monitoring are profound and multifaceted. As organizations increasingly deploy ML algorithms to predict future process behaviors, the need for these systems to be transparent and comprehensible becomes paramount. XAI bridges the gap between the complexity of ML models and the operational necessity for clarity and accountability in decision-making processes. In industries where process outcomes are critical, such as healthcare, the ability of stakeholders to understand and trust AI-based predictions is not a luxury but a requirement. The practical deployment of XAI in these settings implies that operators and decision-makers can glean insights into the reasoning behind predictions, facilitating informed interventions and strategic planning. For instance, in a manufacturing plant, an interpretable model can illuminate the factors leading to potential equipment failure, enabling preemptive maintenance and reducing downtime \cite{mehdiyev2022deep}. 

Furthermore, the practicality of explainability extends to the adaptability
and scalability of interpretability methods. In the ever-changing landscape
of process data, AI systems must provide timely and contextually relevant
explanations. The need for explanations to be customizable and aligned with users’ varying levels of expertise and objectives. This adaptability ensures that AI serves its intended purpose effectively across different contexts and user groups, a critical consideration in business process management's diverse and dynamic environments. 

Moreover, XAI can play a pivotal role in regulatory compliance and risk management. In sectors like finance or law, where predictive models are used to make significant decisions, regulators increasingly demand transparency. XAI methods that can elucidate the logic behind loan application processes or patient pathway assessments are beneficial and may soon be mandated as standard practice.

However, translating XAI from theory to practice also may several complexities. One of the primary concerns is the integration of XAI systems within existing IT infrastructures. Many organizations operate on legacy systems, and introducing sophisticated XAI solutions requires careful planning and execution to ensure compatibility and minimal disruption to ongoing operations. Another practical implication is the need for user training and adaptation. The effectiveness of an XAI system is contingent on the end-user's ability to interpret and act upon the explanations provided. This necessitates training programs to enhance the AI literacy of the workforce, ensuring that users can leverage the full potential of XAI in their day-to-day responsibilities. Furthermore, the economic impact of implementing XAI systems must be considered. Organizations need to evaluate the cost-benefit ratio of adopting such technologies, weighing the potential savings from improved process efficiencies against the investment in technology and training. The practical implications of XAI also extend to the continuous monitoring and updating of these systems. As processes evolve and new data becomes available, XAI models must be maintained and retrained to ensure their explanations remain accurate and relevant. This ongoing maintenance requires a commitment to resource allocation and a strategy for long-term management.

In conclusion, the practical implications present a complex array of challenges and opportunities. For XAI to be successfully integrated into predictive process monitoring, organizations must navigate the technical, operational, and economic landscapes, balancing the promise of AI-driven insights with the realities of their application in the real world. As the field of XAI matures, this pragmatic approach will likely dictate the success and proliferation of explainable systems in industry.

\subsection{Scientific and Theoretical Implications}
The integration of XAI within predictive process monitoring is not just a practical enhancement; it represents a paradigm shift in how scientific inquiry and theoretical development are approached in the context of complex systems.

From a scientific perspective, the incorporation of XAI opens new avenues for research in algorithmic transparency and interpretability. It challenges the conventional black-box approach to ML, calling for novel algorithms and models that are inherently interpretable or can be paired with explanation mechanisms. This need accelerates advancements in areas like feature importance analysis, counterfactual explanations, and causal inference models, all of which contribute to a deeper understanding of the underlying mechanics of complex predictive models. For instance, recently, novel approaches in Uncertainty Quantification (UQ) for predictive process monitoring have been proposed, which are crucial for the way we understand and interact with AI models \cite{weytjens2022learning, mehdiyev2023quantifying}. These innovative methodologies are enhancing transparency by providing insights into the confidence levels of model predictions and even generated explanations \cite{mehdiyev2023communicating}. This shift marks a significant stride towards more transparent, reliable, and user-centric AI systems.

In the theoretical realm, XAI stimulates a re-evaluation of existing theories related to decision-making, cognition, and information processing. It brings to light questions about the nature of understanding and trust in automated systems. For instance, what constitutes a "good" explanation in a predictive process monitoring context, and how do these explanations impact human decision-making and trust? The pursuit of answers to these questions encourages interdisciplinary collaboration, drawing from fields such as psychology, cognitive science, and philosophy to enrich the theoretical underpinnings of XAI.

Furthermore, XAI's focus on interpretability and explainability mandates a rigorous theoretical understanding of the processes being monitored. This requirement not only reinforces the need for domain expertise in model development but also promotes a more symbiotic relationship between domain experts and data scientists. In this context, predictive process monitoring becomes a collaborative scientific endeavor, blending empirical data analysis with domain-specific insights to produce models that are both high-performing and understandable.

The scientific implications of XAI also extend to the validation and evaluation of AI models. Traditional performance metrics like accuracy, precision, and recall are no longer sufficient. XAI introduces the need for new metrics and methodologies that can assess the quality of explanations in terms of relevance, completeness, and comprehensibility. This evolution reflects a broader shift in the scientific community's approach to evaluating AI, placing equal emphasis on the interpretability and operational effectiveness of the models.

From a theoretical standpoint, XAI challenges and refines our understanding of concepts like causality, uncertainty, and prediction. It encourages a more nuanced exploration of how these elements interplay in complex systems and how they can be effectively communicated to users. This exploration has profound implications for theoretical models across various domains, from supply chain management to healthcare, where understanding the causal relationships and uncertainties inherent in predictive models is crucial for effective decision-making.

In summary, the integration of XAI in predictive process monitoring is catalyzing significant scientific and theoretical advancements. It is driving the development of new algorithms and models, fostering interdisciplinary research, redefining evaluation methodologies, and deepening our understanding of complex systems. As the field progresses, the continued exploration of these scientific and theoretical implications will be instrumental in realizing the full potential of XAI, not only as a tool for enhanced predictive analytics but also as a beacon for responsible and transparent AI development.

\section{Conclusion}\label{sec8}
In conclusion, our systematic literature review (SLR), guided by the PRISMA framework, has critically examined the landscape of explainable and interpretable ML within the specialized domain of predictive process mining. By distinguishing between intrinsically interpretable models and more complex black-box models requiring post-hoc explanation, our research has navigated through the multifaceted intricacies of AI and ML systems. Our analysis has not only underscored the practical and academic necessity of explainability and interpretability in building user trust and understanding but also highlighted the specific challenges and opportunities within process mining. As we look forward, the path to fully interpretable and explainable predictive process monitoring is both promising and fraught with challenges. The evolving nature of ML methods and the increasing complexity of data patterns demand continuous and rigorous research. For practitioners and researchers alike, our study serves as a beacon, illuminating the current state of the field and providing a structured foundation for future inquiry and application. It is our hope that this work will inspire further innovation and collaboration, advancing us towards a future where intelligent systems are not only powerful but also transparent, trustworthy, and aligned with human values and understanding.


\end{document}